\documentclass[times, final, 10pt]{elsarticle}

\usepackage{amssymb}
\usepackage{amsmath}
\usepackage{booktabs} 
\usepackage{multirow} 
\usepackage{hyperref}
\usepackage{tabularx}

\begin{document}

\begin{frontmatter}

\title{Temporal Preservation over Processing: Diagnosing and Designing Spatiotemporal Single-Stage Video Detectors\tnoteref{Temporal Preservation over Processing in Single-Stage Video Detectors}}     

\author[first]{Karam Tomotaki-Dawoud\corref{cor1}}
\ead{karam.tomotaki-dawoud@hhi.fraunhofer.de}
\author[first]{Anna Hilsmann}
\ead{anna.hilsmann@hhi.fraunhofer.de}
\author[first]{Peter Eisert}
\ead{peter.eisert@hhi.fraunhofer.de}
\author[first]{Sebastian Bosse}
\ead{sebastian.bosse@hhi.fraunhofer.de}

\affiliation[first]{organization={Fraunhofer HHI}, city={Berlin}, country={Germany}}

\cortext[cor1]{Corresponding author}

\begin{abstract}
Single-stage video object detectors are increasingly deployed in time-critical applications, yet it remains unclear whether these models genuinely reason over temporal context or merely exploit a single informative frame—a gap hidden by standard metrics, which reward correct predictions regardless of how they are reached. We address this from two complementary directions: first, we propose TemporalLens, a model-agnostic diagnostic framework probing temporal dependence through controlled perturbations, structured occlusions, temporal shuffling, redundancy injection, and resolution degradation, revealing whether a detector actually uses information across time. Applied to stacked-frame 2D detectors and our YOLO-3D architecture, it exposes behavioural differences invisible to mAP: stacked 2D models collapse when the target frame is removed, while spatiotemporal models recover predictions from earlier frames, a signature of real temporal reliance. Second, we detail YOLO-3D, a modular real-time spatiotemporal detector built on YOLOv8, and show that simply preserving temporal depth through the backbone is the dominant performance driver (+3.7 pp mAP@50 at 32 frames averaged across scales). Together, the diagnostics and architecture turn "does this detector reason over time?" into a measurable, actionable question.
\end{abstract}

\begin{highlights}
\item Perturbation suite probes temporal reasoning in single-stage video detectors 
\item  Controlled occlusions expose target-frame reliance invisible to standard mAP 
\item  3D backbone recovers predictions from earlier frames where stacked-2D collapses 
\item Temporal-preserving backbone yields +3.7 pp mAP@50 over standard downsampling 
\item  Framework validated on surgical, agricultural, and action-detection domains
\end{highlights}


\end{frontmatter}



\section{Introduction}
\label{sec:intro}

Many vision tasks are inherently temporal: information unfolds across frames, providing cues that single images cannot capture. Applications such as action recognition, video object detection, and articulated pose estimation rely on temporal context to handle occlusions, motion blur, and rapid appearance changes. Modern single-stage video detectors process multiple frames jointly and report consistent accuracy gains over their single-frame counterparts. However, it remains unclear to what extent these models actually exploit information across time. Do they learn meaningful temporal representations, or do they primarily base their predictions on a single highly informative frame? Answering this question requires both (i) an architecture with explicit, controllable temporal operators whose contribution can be isolated, and (ii) a diagnostic methodology that can expose single-frame dependence hidden behind headline accuracy metrics.

\noindent To this end, we make two complementary contributions: \textbf{YOLO-3D}, a lightweight spatiotemporal model built on YOLOv8~\cite{yolov8_ultralytics}, whose temporal operators can be isolated and stress-tested, and a \textbf{model-agnostic diagnostic framework} that applies controlled perturbations---occlusion schedules, temporal shuffling, redundancy injection, and mid-sequence resolution degradation---to directly measure temporal integration.

Throughout this work, we distinguish between \emph{stacked-frame 2D models}, which extend the YOLO family by treating multiple frames as extended input channels and applying purely spatial 2D convolutions without temporal locality or order, and \emph{3D spatiotemporal models}, which use convolutions that explicitly span the time dimension via local temporal kernels, thereby enforcing temporal continuity. This distinction is central to our analysis: the proposed diagnostics determine whether explicit temporal operators translate into meaningful sequence integration, while the ablation study identifies \emph{which} architectural decisions enable that integration.

A key insight from our ablation is that temporal reasoning in the neck and head is bottlenecked by the backbone's temporal downsampling schedule. Conventional 3D backbones aggressively reduce the temporal dimension, often collapsing it to $D{=}1$ before features reach the neck, rendering any downstream temporal attention ineffective on short sequences. Our temporal-preserving backbone addresses this by maintaining uniform temporal depth across mid-to-late pyramid levels; on UCF101-24 this single design choice yields the largest accuracy gains ($+3.7$\,pp mAP@50 at 32 frames averaged across scales), outweighing more complex fusion or squeeze modules and establishing that \emph{preserving} temporal information matters more than sophisticated temporal \emph{processing}.

Complementing the architectural study, our diagnostic framework reveals \emph{behavioral} differences that standard mAP metrics obscure. Applied across surgical scene analysis (kidney transplantation) and articulated animal pose estimation (dairy-cow skeleton), it exposes clear failure modes: stacked-2D models collapse catastrophically when the target frame is removed, whereas even the vanilla 3D variant retains predictive capability by integrating earlier frames, while conversely exhibiting heightened sensitivity to mid-sequence spatial degradation, confirming genuine rather than incidental temporal reliance. The same probes show that temporal reasoning provides limited benefit in appearance-dominated surgical scenes but substantially improves robustness under occlusion and pose ambiguity in motion-dependent pose estimation.

Our contributions are twofold.

\begin{figure}
    \centering
    \includegraphics[width=0.75\linewidth]{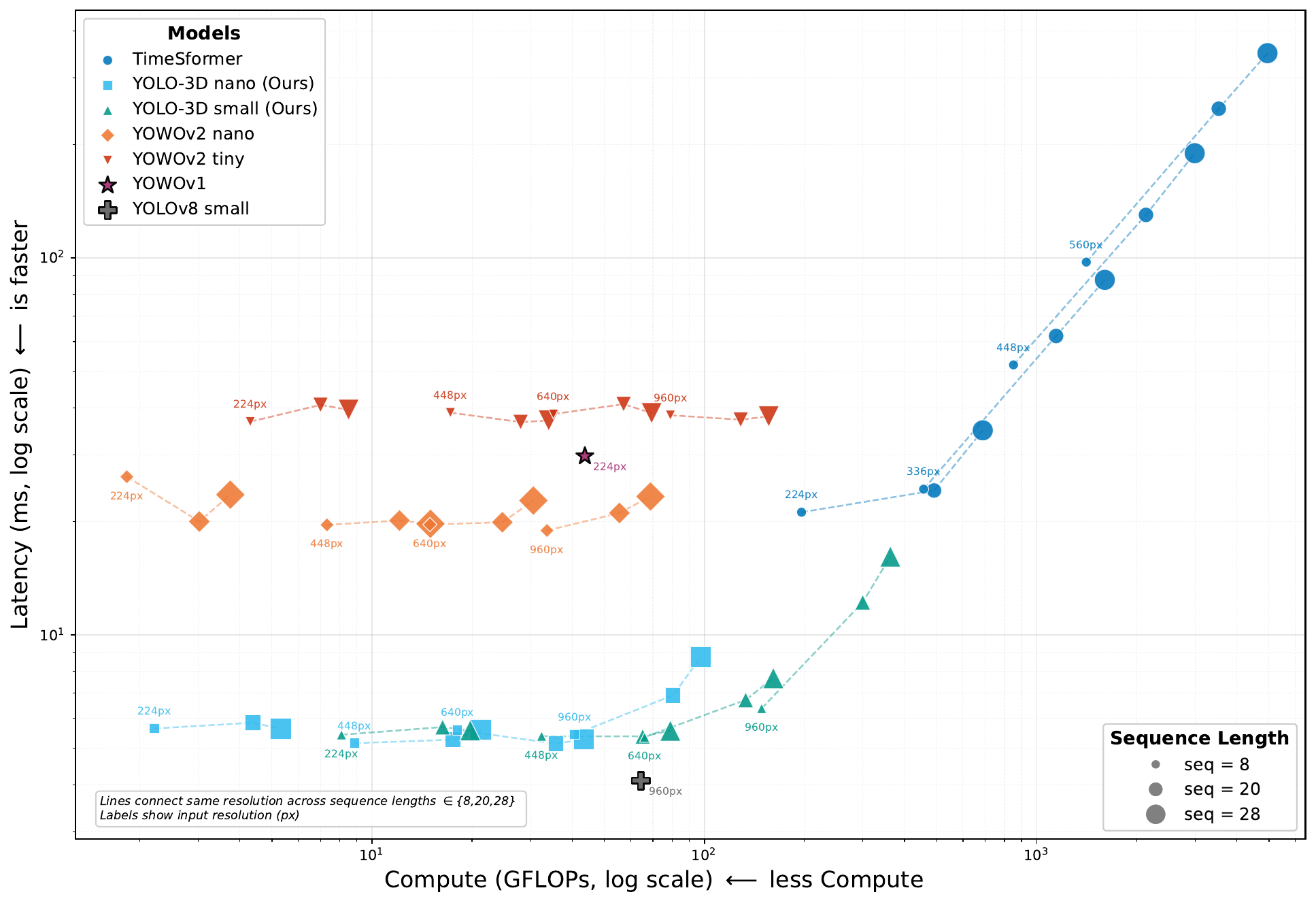}
    \caption{
    Latency–compute across spatiotemporal action detection architectures. Marker shape and color distinguish models; marker size encodes sequence length $\in \{8,20,28\}$; lines connect configurations sharing the same resolution. Our YOLO-3D vanilla (nano/small variants) occupies the low-compute/low-latency regime (2–361 GFLOPs, 5–21 ms), maintaining sub-25 ms inference even at 960p with 28 frames. YOWOv2 variants exhibit resolution-invariant latency ($\approx19$-$42$ ms) but scale compute with input size. TimeSformer incurs 1–2 orders of magnitude higher compute and latency, reaching 4.9 TFLOPs and 450 ms at $560p\times28$ frames. Reference points include YOWOv1 (224p, T= 16) and single-frame YOLO-v8s baseline at 960p. All measurements in FP32 on an RTX 4090}
    \label{fig:latency_tradeoff}
\end{figure}

\textbf{1. A modular spatiotemporal architecture for real-time video detection} We build on YOLOv8 with (a)~a temporal-preserving 3D backbone that retains non-degenerate temporal depth at all FPN levels, (b)~Spatio-Temporal Attention Fusion (STAF) with factored linear attention for learned cross-scale fusion, and (c)~Adaptive Temporal Focus (ATF) for key-frame-aware temporal squeeze. The architecture maintains real-time performance (Fig.~\ref{fig:latency_tradeoff}) across nano, small, and medium scales.

\textbf{2. A model-agnostic diagnostic framework for temporal reasoning} We propose a reusable perturbation suite that systematically stress-tests whether video detectors rely on temporal information or collapse to single-frame cues, enabling principled model selection beyond headline mAP.

We validate these contributions through (i) a comprehensive architectural ablation on UCF101-24 (10 configurations $\times$ 3 model scales $\times$ 2 sequence lengths $=$ 60 experiments) that factorizes the design space into backbone type, neck fusion, and temporal squeeze strategy; and (ii) diagnostic evaluation on the two real-world domains above. The division of labor is deliberate: the perturbation diagnostics isolate the backbone-level temporal contrast (vanilla 3D vs.\ stacked-2D) on the real-world datasets, while the attention-fusion and temporal-focus modules are evaluated through the controlled UCF101-24 ablation.


\section{Related Work}
\label{sec:RelatedWork}

\paragraph{Video Models}

Early deep video architectures extended 2D CNNs into the temporal domain using 3D convolutions. C3D~\cite{Tran2015C3D} demonstrated that spatiotemporal kernels can jointly capture appearance and motion, while I3D~\cite{Carreira2017I3D} leveraged 2D pretrained networks for effective transfer learning to action recognition. Subsequent developments explored efficiency–accuracy trade-offs: 3D ResNets~\cite{Hara2018Res3D}, factorized R(2+1)D~\cite{Tran2018Closer}, dual-rate SlowFast networks~\cite{Feichtenhofer2019SlowFast}, and X3D~\cite{Feichtenhofer2020X3D}, which systematically scales network dimensions. Transformer-based video models further advanced temporal modeling through attention mechanisms. TimeSformer~\cite{Bertasius2021TimeSformer}, ViViT~\cite{Arnab2021ViViT}, and Video Swin~\cite{Liu2022VideoSwin} achieve strong accuracy in classification tasks, but at high computational cost. While recent self-supervised approaches such as VideoMAEv1-v2~\cite{Tong2022VideoMAE,Wang23VideoMAEV2} and VideoMAEv2~\cite{Wang23VideoMAEV2} improve data efficiency, yet still prioritize classification and operate at reduced input sizes. Collectively, these works demonstrate powerful temporal reasoning but remain prohibitive for real-time deployment in detection/pose tasks.

\paragraph{Temporal Extensions of Single-Stage Detectors}
The YOLO family has evolved toward real-time efficiency~\cite{Redmon2018YOLOv3,Bochkovskiy2020YOLOv4}, but remains fundamentally frame-based. Several works introduce temporal context: YOWO~\cite{Kopuklu2021YOWO} couples a 2D and a 3D CNN branch with a YOLO-based detection head for real-time spatiotemporal action localization; YOWOv2~\cite{Zhiqiang2025YOWOv2} replaces YOWO's single-level anchor-based design with a multi-level FPN for the 2D backbone branch and anchor-free heads, and introduces a channel encoder that fuses the upsampled 3D spatiotemporal features with each 2D FPN level through channel concatenation followed by a channel self-attention mechanism, improving action accuracy at real-time speed;
YOLOV~\cite{Shi2023YOLOV} aggregates short-range temporal features with modest overhead; and frame-stacking strategies such as Temporal-YOLO~\cite{Corsel_2023_WACV}, Temporal-YOLOv8~\cite{vanLeeuwen2024TemporalYOLOv8}, and TYOLOv8~\cite{van_Lier_2025_WACV} encode consecutive frames as extended input channels.
These methods improve accuracy in specific settings, but they differ from our work in two important respects. First, they generally treat the backbone's temporal downsampling schedule as inherited from the classification literature, whereas we identify aggressive temporal striding as the dominant bottleneck and show that a simple temporal-preserving strategy outweighs more complex neck or head modifications. Second, none of them offers a principled evaluation of whether their temporal extensions genuinely integrate sequence information or merely benefit from augmented input bandwidth.
Beyond the YOLO family, other real-time single-stage detectors inject temporal context through explicit motion streams: Zhang et al.~\cite{ZHANG2020LearningMotion} jointly learn action localization and optical-flow estimation end-to-end with multi-scale appearance--motion fusion, while ACDnet~\cite{LIU2021ACDnet} augments an SSD detector with flow-guided feature approximation and memory aggregation from 3 frames, sustaining real-time speed and retaining decent accuracy. Efficiency-driven VOD approaches such as MaskVD~\cite{Sarkar_MaskVD2025} and FAIM~\cite{Hashmi_2025_FAIM} reuse features or masks across frames, reducing compute without directly assessing temporal dependence.

\paragraph{Perturbation and Occlusion Analyses}
Perturbation-based explanations offer insights into spatiotemporal feature usage. Adaptive occlusion-sensitivity analysis for video CNNs~\cite{Tomoki_AOSA23} identifies influential regions over time, while classical occlusion~\cite{Zeiler_occlusion} and its generative extensions~\cite{Agarwal_Occ2020} reveal feature dependence in static images. Broader structured perturbation work~\cite{NEURIPS2019_Hooker,Samek2021,dawoud2023} highlights model vulnerabilities under masking. These methods provide \emph{post-hoc interpretability} of individual predictions but do not offer a systematic way to quantify a detector's reliance on temporal context as a whole. Our diagnostic framework closes this gap: rather than explaining which spatial or temporal regions matter for a single sample, it systematically stress-tests whether the architecture integrates information across time, exposing failure modes that headline mAP scores obscure.

\paragraph{Summary}
Prior video models, YOLO-style temporal extensions, and perturbation-based analyses advance temporal modeling or interpretation individually, but none combines (i)~an architectural analysis that pinpoints the backbone's temporal downsampling as the key design lever with (ii)~a controlled diagnostic protocol that verifies whether temporal operators translate into genuine sequence integration. Our work addresses both gaps jointly: YOLO-3D provides a modular architecture whose components can be isolated and ablated, while the proposed perturbation suite offers a reusable evaluation instrument applicable to any video detector.

\section{YOLO-3D}
\label{sec:ModelArchitecture}

We build on the YOLOv8~\cite{yolov8_ultralytics} design, lifting its key spatial operations to lightweight 3D counterparts. The design preserves YOLOv8's two-part structure (backbone + Neck/head) while enabling temporal feature extraction with minimal changes to the original architecture. The resulting system, YOLO-3D, is organised around three axes of design: the 3D backbone and its temporal downsampling schedule (Sec.~\ref{sec:backbone3d}), the neck fusion strategy that aggregates features across pyramid levels (Sec.~\ref{sec:staf}), and the temporal squeeze that collapses the 3D feature volume to a 2D map for the detection head (Sec.~\ref{sec:atf}). Along each axis we provide a simple, parameter-free \emph{vanilla} option as well as a learned, attention-driven alternative, enabling systematic ablation of component contributions (Sec.~\ref{sec:ablation-ucf}).

\begin{figure}
  \centering
  \includegraphics[width=0.95\textwidth]{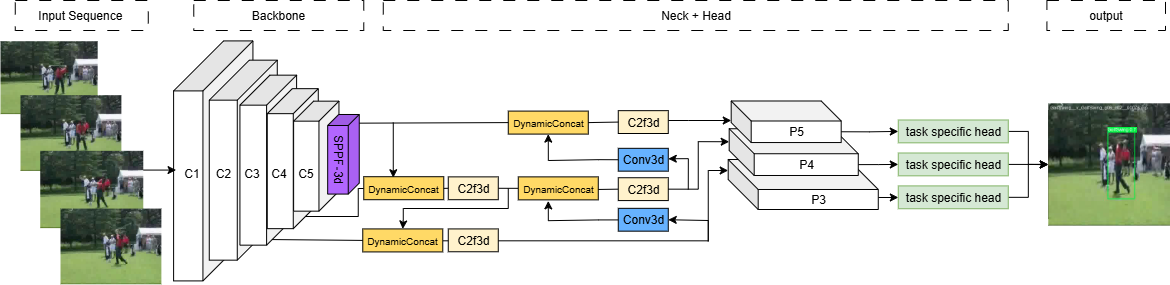}
  \includegraphics[width=0.9\textwidth]{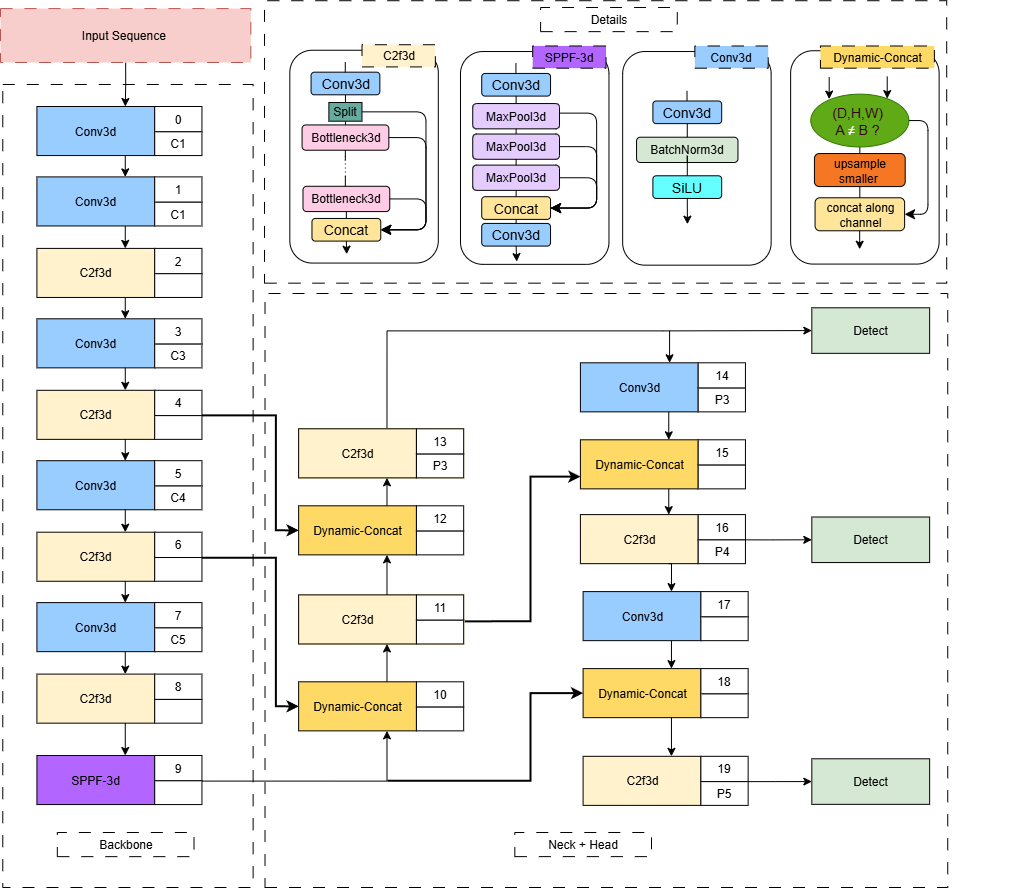}
  \caption{Spatiotemporal extension of YOLOv8. A $T$-frame video clip is processed by a 3D backbone (Conv3d/C2f3d blocks and SPPF3d), followed by 3D FPN/PAN-style fusion. The \emph{vanilla} variant uses DynamicConcat for temporal alignment and AdaptiveAvgPool3d for depth reduction; the \emph{enhanced} variant replaces these with spatio-temporal attention fusion and adaptive temporal focus modules (Sec.~\ref{sec:enhanced-extension}). All variants feed into the standard anchor-free YOLOv8 detection head}
  \label{fig:spatiotemporal-arch}
\end{figure}

\paragraph{Base YOLOv8 Overview}
YOLOv8 consists of a CSPDarknet~\cite{Wang2020CSPNet, Bochkovskiy2020YOLOv4}-inspired backbone with C2f (Cross Stage Partial with two convolutions) blocks and an SPPF (Spatial Pyramid Pooling-Fast) layer, followed by a unified head that performs FPN/PAN-style multi-scale fusion, i.e., top-down and bottom-up aggregation with up/down-sampling and concatenation paths~\cite{Lin2017FPN, Liu2018PANet}, and an anchor-free, decoupled prediction head. All components prioritize real-time efficiency.

\subsection{3D Backbone and Vanilla Spatiotemporal Extension}
\label{sec:backbone3d}

We generalize the YOLOv8 backbone to the temporal domain, as shown in Fig.~\ref{fig:spatiotemporal-arch}, by replacing 2D convolutions with \textbf{Conv3d} layers and extending C2f into \textbf{C2f3d}, which preserves its split-transform-merge structure while learning short-range motion cues. The \textbf{SPPF3d} module applies multi-kernel pooling in space and time to capture broader temporal context. Successive stride-2 Conv3d layers progressively reduce the temporal depth, yielding a multi-scale feature pyramid $\{P_3, P_4, P_5\}$ where deeper levels may have a temporal depth as low as $D{=}1$.

The neck and squeeze stages mirror the FPN/PAN organization of YOLOv8, with all operations lifted to 3D. Cross-scale fusion introduces mismatched temporal depths due to stride differences; we address this with \textbf{DynamicConcat}, which inspects the temporal dimensions of two input feature maps and upsamples the shorter one via nearest-neighbor interpolation along the time axis before concatenation. This prevents fusion failures caused by unequal temporal lengths and ensures consistent cross-scale aggregation. After FPN/PAN fusion, the 3D feature maps are collapsed to 2D by applying \textbf{AdaptiveAvgPool3d} with a target depth of~$1$, producing a uniform temporal average at each spatial location. While simple and parameter-free, this strategy treats all frames equally and discards temporal ordering—an acceptable trade-off when the detection target coincides with the dominant frame in the clip, but a potential limitation when precise temporal localization is required. However, for almost all sequences shorter than 16 frames, the temporal depth is only 1-2. Together, the 3D backbone with DynamicConcat and AdaptiveAvgPool3d constitute the \emph{vanilla} YOLO-3D variant; the detection head remains unchanged from YOLOv8.

\subsection{Temporal Preserving and Attention Fusion Extension}
\label{sec:enhanced-extension}

We replace both the concatenation-based fusion and the average-pool temporal reduction with learned, attention-driven modules. At the same time, the most important factor is changing some of the backbone temporal strides, in order to allow more temporal information to flow to the neck, while the detection head remains unchanged.

\subsubsection{Temporal Resolution Preservation in the Backbone}
\label{sec:temporal-backbone}

Conventional 3D CNN backbones inherit spatial downsampling schedules from their 2D counterparts, applying strided spatiotemporal convolutions at mid-to-late stages (e.g., \texttt{conv3}, \texttt{conv4}, \texttt{conv5}) with joint temporal and spatial strides~\cite{Hara2018Res3D}.
While such temporal downsampling reduces computational cost and expands the temporal receptive field, it rapidly collapses the temporal dimension to a point where downstream temporal reasoning modules, such as attention-based fusion or learned temporal pooling, operate on degenerate single-frame features and are effectively reduced to no-ops.

\paragraph{Quantifying the problem}
Table~\ref{tab:temporal-depth} traces the temporal depth at each pyramid level under both strategies for the range of clip lengths commonly used in action detection benchmarks.
With full temporal downsampling (stride $2{\times}2{\times}2$ at all five backbone convolutions), the temporal dimension is divided by $2^5 = 32$ from input to P5.
At a common training length (8-frame clip), all three pyramid levels collapse to $D{=}1$, rendering any temporal fusion or attention in the neck entirely inactive.
Even at 16 frames, only P3 retains $D{=}2$, while P4 and P5 remain at $D{=}1$.
Across the five sequence lengths examined ($T \in \{8, 12, 16, 24, 32\}$), P5 never exceeds $D{=}1$ under the baseline strategy, and P4 reaches $D{>}1$ only for $T{\geq}24$.
In practice, this means that the spatiotemporal attention modules in the neck, which are designed to exploit temporal context for more discriminative feature fusion, receive at most a single temporal slice and collapse to purely spatial operations.

\begin{table}[t]
\centering
\caption{Temporal depth at each FPN pyramid level under baseline and temporal-preserving backbone strategies for common input clip lengths $T$. Baseline applies temporal stride~2 at all five backbone convolutions; temporal-preserving restricts temporal stride~2 to the two stem layers only}
\label{tab:temporal-depth}
\small
\setlength{\tabcolsep}{3pt}
\begin{tabular}{c ccc ccc}
\toprule
 & \multicolumn{3}{c}{Vanilla baseline ($s_t{=}2$ throughout)} & \multicolumn{3}{c}{Temporal-preserving ($s_t{=}1$ at layers 3,5,7)} \\
$T$ & P3 & P4 & P5 & P3 & P4 & P5 \\
\midrule
8  & 1 & 1 & 1 & 2 & 2 & 2 \\
12 & 2 & 1 & 1 & 3 & 3 & 3 \\
16 & 2 & 1 & 1 & 4 & 4 & 4 \\
24 & 3 & 2 & 1 & 6 & 6 & 6 \\
32 & 4 & 2 & 1 & 8 & 8 & 8 \\
\bottomrule
\end{tabular}
\end{table}

\paragraph{Prior evidence for temporal preservation}
Empirical evidence from several recent architectures corroborates the importance of preserving temporal resolution.
Chen~et~al.~\cite{Chen_2021_CVPR} demonstrate that removing temporal pooling from I3D yields a $+1.1\%$ gain on Kinetics-400 and a more pronounced $+6\%$ gain on Something-Something~V2~\cite{Goyal2017something}—a benchmark that explicitly requires fine-grained temporal reasoning, bringing I3D to parity with temporally-aware models such as TAM~\cite{Liu_2021_ICCV}.
The Fast pathway of SlowFast~\cite{Feichtenhofer2019SlowFast} deliberately omits all temporal downsampling layers throughout the network hierarchy, maintaining the full temporal resolution of feature tensors up to the final global pooling stage; the authors note that temporal downsampling would be detrimental precisely because fine temporal detail is the pathway's primary representational objective.
Aligned with this, the DiST architecture~\cite{Qing_2023_ICCV} explicitly avoids temporal downsampling within its lightweight temporal encoder, preserving temporal detail while delegating spatial compression to a separate branch.

\paragraph{Our approach}
Motivated by both the quantitative analysis above and the prior evidence, our backbone adopts a \textbf{temporal-preserving} downsampling strategy.
We retain temporal stride~2 in the two stem convolutions (layers~0 and~1), which reduce the temporal dimension from $T$ to $T/4$, a manageable compression that keeps early-stage activations within memory budgets.
At the three subsequent strided convolutions (layers~3, 5, and~7, which feed P3, P4, and P5 respectively), we replace the $3{\times}3{\times}3$ kernels with $1{\times}3{\times}3$ kernels and the $(2,2,2)$ strides with $(1,2,2)$, applying spatial downsampling while leaving the temporal dimension intact.
As shown in the right half of Table~\ref{tab:temporal-depth}, this ensures that \emph{all} pyramid levels carry the same temporal depth ($T/4$) throughout the FPN, enabling the spatiotemporal modules in the neck to operate on non-degenerate temporal sequences at every scale.

\paragraph{Cost analysis}
Temporal preservation trades parameters for activation memory at essentially unchanged latency. Replacing the $3{\times}3{\times}3$ kernels with $1{\times}3{\times}3$ at three backbone layers reduces total model parameters (by up to $14.1\%$ when the neck is otherwise unchanged), while retaining $D{>}1$ through the neck raises per-sample
activation memory by $30$--$46\%$ depending on clip length---the primary cost, and the only one that may require modest batch-size adjustment on memory-constrained hardware. This overhead is constant in relative terms across the $n$/$s$/$m$ scales, making the trade-off predictable when
scaling the architecture. ~\ref{app:cost} gives the full per-dimension breakdown and measurements (Table~\ref{tab:cost-tradeoff}).

\begin{figure}
  \centering
  \includegraphics[width=0.75\textwidth]{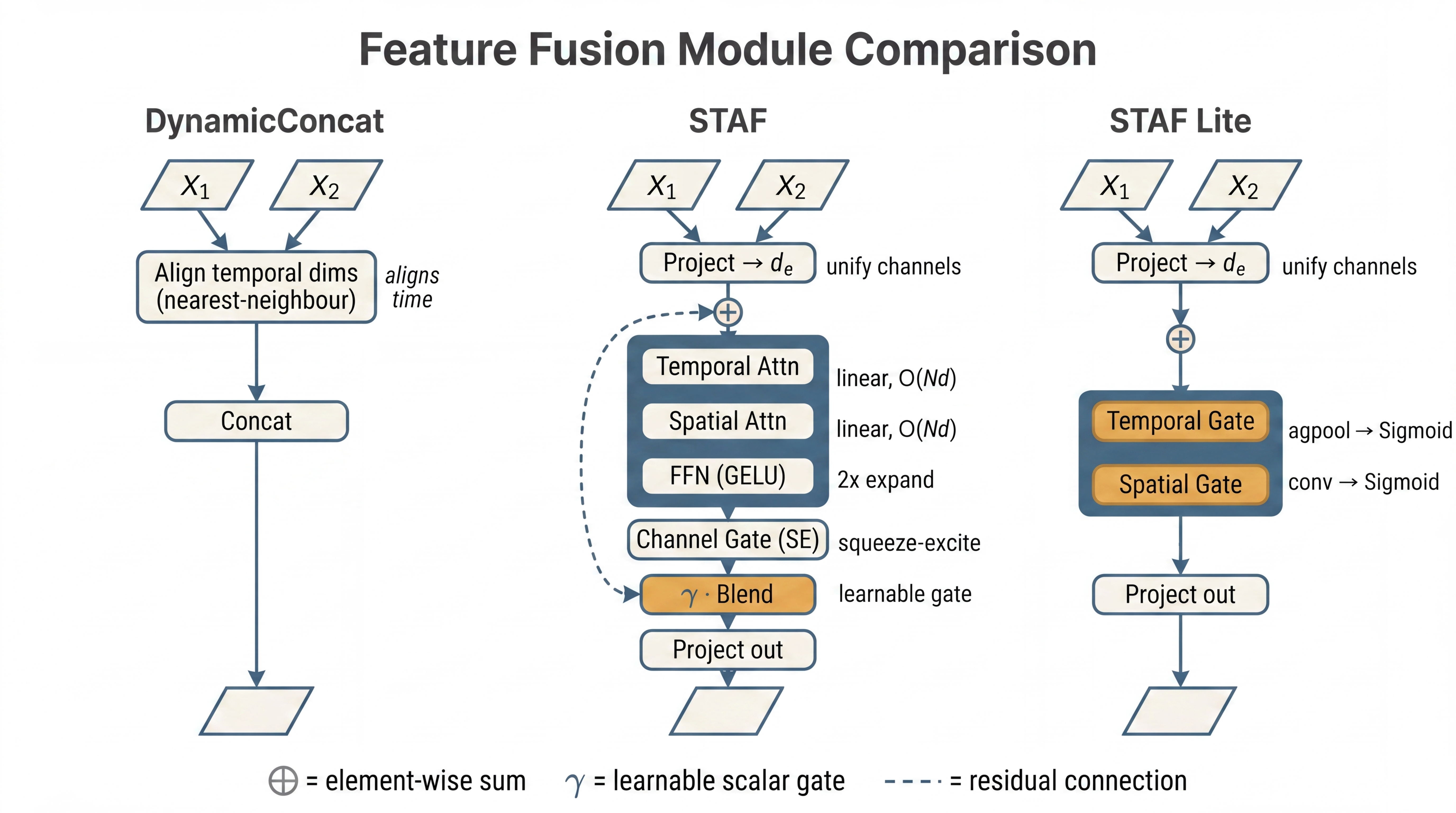}
  \caption{
  Architectural comparison of the three cross-scale feature fusion modules: (Left)~\textbf{DynamicConcat} resolves temporal mismatches between adjacent feature pyramid levels via nearest-neighbor interpolation before concatenation.(Center)~\textbf{STAF (SpatioTemporalAttentionFusion)} projects both inputs to a shared embedding dimension $d_e$, fuses them via element-wise summation, and refines the result through factored temporal and spatial linear attention ($\mathcal{O}(Nd)$ complexity) followed by a squeeze-and-excitation channel gate~\cite{Hu2018SENet}; a learnable scalar $\gamma$ controls the residual blend before the final projection. (Right)~\textbf{STAFLite} retains the same projection, alignment, and gated residual structure as STAF but replaces the attention blocks with lightweight average-pooling temporal gating and convolutional spatial gating, reducing parameter count while preserving the learned fusion benefits.
  }
  \label{fig:spatiotemporal-Fusion-module}
\end{figure}

\subsubsection{Spatio-Temporal Attention Fusion}
\label{sec:staf}

We replace DynamicConcat with \textbf{SpatioTemporalAttentionFusion} (STAF), an attention-based cross-scale fusion module. 

\noindent Given two feature maps $\mathbf{x}_1 \in \mathbb{R}^{B \times C_1 \times D_1 \times H_1 \times W_1}$ and $\mathbf{x}_2 \in \mathbb{R}^{B \times C_2 \times D_2 \times H_2 \times W_2}$ from adjacent FPN levels, STAF proceeds in five stages, shown in Fig.~\ref{fig:spatiotemporal-Fusion-module}:

\textbf{1. Projection} Each input is mapped to a shared embedding dimension $d_e = \min(C_1, C_2)$ via a $1{\times}1{\times}1$ Conv3d--BatchNorm--SiLU block.
\textbf{2. Alignment} Spatial and temporal dimensions are matched via trilinear interpolation to the element-wise maximum of each axis.
\textbf{3. Factored spatiotemporal attention} The element-wise sum of the two projections is processed by a factored attention block that decomposes the 3D volume into:
      \emph{A) Temporal attention} across $D$ for each spatial location, and
      \emph{B) Spatial attention} across $H{\times}W$ for each temporal slice,
    each implemented as kernel-linearised multi-head attention~\cite{Katharopoulos2020Transformers} with $\mathrm{ELU}{+}1$ feature maps, yielding $O(N\,d)$ complexity instead of $O(N^2)$. When $D{=}1$, temporal attention is skipped entirely. A position-wise feed-forward network (two linear layers with a GELU non-linearity and $2{\times}$ expansion) follows the two attention stages. The design draws on factored attention strategies employed in efficient vision transformers~\cite{zheng2023efficient, zheng2022linear} and the channel/spatial self-attention modules of YOWOv2~\cite{Zhiqiang2025YOWOv2}.
\textbf{4. Channel gating} A squeeze-and-excitation gate~\cite{Hu2018SENet} modulates the attended features channel-wise.
\textbf{5. Gated residual and output projection} The attended features are blended into the fused representation via a learnable scalar $\gamma$ (bounded by sigmoid), followed by a $1{\times}1{\times}1$ projection to the target output channels.

\noindent Numerical stability under mixed-precision training is ensured by performing all attention arithmetic in float32 with $N$-aware key scaling and a clamped normalizer, following best practices for linear attention in low-precision regimes~\cite{zheng2023efficient}. Per-stage learnable residual scales bound the contribution of each attention and FFN branch.

\paragraph{Lite variant}
For latency-sensitive deployments we also provide \textbf{SpatioTemporalAttentionFusionLite}, which preserves the project-align-fuse structure but replaces the factored self-attention with lightweight \emph{multiplicative} gating. Both inputs are projected by $1{\times}1{\times}1$ convolutions, aligned by nearest-neighbor interpolation, summed, and passed through a BatchNorm-SiLU block; the result is then refined by two sequential gates:
\begin{itemize}
  \item A \emph{temporal gate} applies adaptive average pooling along the depth axis followed by a $1{\times}1{\times}1$ convolution and sigmoid activation, producing per-channel temporal importance weights.
  \item A \emph{spatial gate} applies a $1{\times}1{\times}1$ convolution to a single-channel map and sigmoid activation, highlighting salient spatial regions.
\end{itemize}
Both gates modulate the fused features by direct element-wise multiplication. Unlike STAF, the Lite variant uses neither the factored attention block nor a learnable scalar residual blend ($\gamma$): the gated features are passed straight to the output projection. It thereby incurs roughly $40\%$ fewer parameters than the full STAF while retaining the learned cross-scale fusion and channel/spatial reweighting. Both variants accept their input channel counts automatically from the model parser, ensuring compatibility with all width/depth scaling factors.

\subsubsection{Adaptive Temporal Focus}
\label{sec:atf}

The average-pool depth squeeze (performed by AdaptiveAvgPool3d) before the prediction head is replaced by \textbf{AdaptiveTemporalFocus} (ATF), which converts the 3D feature map to 2D ($B{\times}C{\times}D{\times}H{\times}W \to B{\times}C{\times}H{\times}W$) while preserving temporal awareness. ATF supports three operating modes:

\textbf{a) Explicit anchor} (\texttt{last} or \texttt{middle}): the feature slice at the designated key-frame index is taken as the primary detection signal.
\textbf{b) Learned selection} (\texttt{learn}): a lightweight $1{\times}1{\times}1$ convolution followed by softmax over the temporal axis produces per-channel frame-attention weights, yielding a soft-selected anchor without requiring prior knowledge of the key-frame position.

\noindent In all modes, a gated context summary is computed in parallel: a sigmoid-activated $1{\times}1{\times}1$ convolution produces per-frame importance weights, which are used to form a weighted temporal average. The final output blends the anchor with the context via a learnable mixing coefficient~$\alpha$ (initialized to $0.1$):
\begin{equation}
  \mathbf{y} = \mathbf{f}_{\mathrm{anchor}} + \alpha \cdot \mathbf{f}_{\mathrm{context}}\,,
  \label{eq:atf}
\end{equation}
where $\mathbf{f}_{\mathrm{anchor}}$ is the selected (hard or soft) key-frame and $\mathbf{f}_{\mathrm{context}}$ is the gated summary. This two-path design provides a stronger inductive bias than a pure attention-weighted average: when the key-frame position is known, the explicit anchor achieves near-perfect reconstruction of the target frame, while the context path supplies complementary motion cues. When the position is unknown, the learned mode degrades gracefully to a slightly-informed average that can be refined end-to-end by the detection loss.

\paragraph{Discussion} Compared to the vanilla extension, the enhanced modules introduce a modest parameter overhead (governed by the shared embedding dimension $d_e$ for STAF and a single $C{\times}C$ convolution for ATF) but offer three practical advantages: (i)~cross-scale fusion is \emph{learned} rather than heuristic, letting the network weight the importance of each feature level; (ii)~temporal aggregation preserves frame ordering and can focus on the detection-relevant key-frame; and (iii)~both modules are scale-agnostic, their channel dimensions are derived automatically from the model's width multiplier, enabling a single configuration file to serve all model sizes (nano through extra-large). The loss functions and the anchor-free detection formulation remain unchanged from YOLOv8.

\section{Diagnostic Framework: TemporalLens}
\label{sec:diagnostic}

One of our goals is to determine whether a video detector genuinely
leverages temporal context or effectively reduces to single-frame
inference. We adopt an occlusion-based perturbation strategy in which
systematically masking portions of the input sequence reveals their
contribution to model predictions. While occlusion directly probes
input-segment influence~\cite{Zeiler_occlusion}, naive black or gray
masking introduces distribution shift~\cite{NEURIPS2019_Hooker};
generative inpainting mitigates this but incurs substantial
overhead~\cite{Agarwal_Occ2020}. Following evidence that simple
occlusion remains competitive with more sophisticated explainability
methods~\cite{Samek2021,dawoud2023}, we replace masked regions with the
dataset-wide RGB mean $\boldsymbol{\mu}$, mitigating distribution shift
at negligible cost.

Since exhaustive masking across all $2^T$ temporal combinations is
infeasible, we design a suite of seven structured perturbations, each
targeting a complementary aspect of temporal reasoning: early-frame
reliance, temporal continuity, frame-order sensitivity, redundancy
dependence, and robustness to spatial degradation---factors identified
as essential to temporal analysis in prior video understanding
research~\cite{Suzuki_2018_ECCVWS,Feichtenhofer2019SlowFast}.
Table~\ref{tab:perturbations} summarizes each probe and the behavioural
signature it is designed to elicit; the formal masking operators, the
generalization to an arbitrary target-frame index $\tau$, and the
per-probe expectations are specified in~\ref{app:perturbations}.

\begin{table}[t]
\centering
\caption{The TemporalLens perturbation suite. Each probe degrades the
input clip along one temporal axis; the rightmost column gives the
behavioural signature expected of a genuinely temporal model. Formal
definitions are given in ~\ref{app:perturbations}}
\label{tab:perturbations}
\footnotesize
\begin{tabularx}{\linewidth}{@{}l l X@{}}
\toprule
Probe & Manipulation & Temporal property / expected signature \\
\midrule
FP-$p$  & Mask first $\lceil pT\rceil$ frames
        & Early-context reliance; accuracy declines as $p$ grows \\
ES-$s$  & Mask alternating frames
        & Continuity; hurts motion integrators, eased by redundancy \\
HL      & Mask final frame
        & Last-frame reliance; 3D recovers from earlier frames \\
CL      & Keep only final frame
        & Upper bound without temporal context \\
TS      & Shuffle all frames
        & Order sensitivity; degrades sequential models \\
\,TS-EL & Shuffle all but final frame
        & Isolates order vs.\ target-frame dominance \\
FR-$k$  & Duplicate every $k$-th frame
        & Novelty dependence; penalises motion-cue models \\
RD-$p$-$q$ & Downsample middle $p\%$ by $q$
        & Spatial-detail robustness vs.\ temporal compensation \\
\bottomrule
\end{tabularx}
\end{table}

\paragraph{Quantifying sensitivity: the $\Delta_X$ metric}
To compare the influence of each perturbation across models and tasks,
we define the diagnostic sensitivity metric $\Delta_X$ as the difference
between perturbed and baseline performance:
\begin{equation}
    \Delta_X = \text{mAP}_X - \text{mAP}_{\text{baseline}},
\end{equation}
where $\text{mAP}_{\text{baseline}}$ is performance on the unperturbed
full sequence and $\text{mAP}_X$ is performance under perturbation $X$.
A negative $\Delta_X$ indicates degradation, with larger magnitudes
reflecting stronger reliance on the probed aspect (e.g.,
$\Delta_{\text{HL}} \ll 0$ implies heavy last-frame dependence), while
$\Delta_X \approx 0$ suggests robustness or irrelevance of that
property. The metric enables direct comparison of architectural
sensitivities: if
$|\Delta_{\text{HL}}^{\text{2D}}| \gg |\Delta_{\text{HL}}^{\text{3D}}|$,
the stacked-2D model exhibits greater last-frame reliance than its 3D
counterpart. We report $\Delta_X$ alongside absolute mAP in all ablation
tables to aid interpretation.

\section{Experiments and Results}
\label{sec:ExpResults}

To quantify the degree to which models exploit temporal information rather than relying on a single frame, we measure performance degradation under controlled perturbation protocols and compare two architectural variants:
\begin{itemize}
    \item \textbf{YOLOv8-2D (stacked temporal):} standard YOLOv8 applied to $(3T,H,W)$ channel-concatenated input, which treats sequences as extended channels and applies 2D convolutions over $(3T)$ channels, suggested by~\cite{vanLeeuwen2024TemporalYOLOv8,van_Lier_2025_WACV} for the YOLOv8 model.
    \item \textbf{YOLO-3D Vanilla (temporal backbone / ours):} YOLOv8 model with a 3D-convolutional backbone operating on $(T,3,H,W)$ input, a 3D-adjusted neck, and task-specific head. Task heads and losses remain unchanged from YOLOv8 to isolate the effect of spatiotemporal feature extraction.
\end{itemize}
 For fairness, each protocol is exported in two serialized formats: \emph{stacked} TIFFs $(3T,H,W)$ for the 2D model (with channel order $R_1,G_1,B_1,\dots,R_T,G_T,B_T$) and \emph{unstacked} NPYs $(3,T,H,W)$ for the 3D model, ensuring bitwise-identical conditions across architectures. Our primary diagnostic experiments target two domain-specific, real-world datasets in surgery and veterinary surveillance:

\begin{enumerate}
    \item \textbf{Operating Room (Object and Action Detection):} 
    Surgical scenes from kidney transplantation with seven action classes (skin incision/stitching, kidney pre/post-reperfusion, surgical bowl, artery forceps, and opened surgical site).
    The dataset consists of 5,492 training and 1,372 validation samples. Each sample is a 12-frame sequence with a 4-frame skip ($\approx 2$ seconds at 30 fps). Challenges include occlusions, variable illumination, and non-rigid motion.
    
    \item \textbf{Dairy-Cow Pose Skeleton (Pose Estimation):} Farm surveillance videos depicting adult dairy cows (Bos taurus) for articulated keypoint detection and pose estimation, a task critical for lameness diagnosis. The dataset contains 1,484 training and 366 validation samples. Each sample is a 20-frame sequence with a skip of 2 frames ($\approx 2$ seconds at 30 fps). Robustness to occlusion is important.

\end{enumerate}

Both datasets are annotated with ground-truth bounding boxes (and 23 keypoints for the dairy cow pose estimation task) of the last frame in each sequence. This design makes Hide-Last and Last-Only particularly diagnostic for last-frame reliance.  Results are reported following COCO’s evaluation protocol~\cite{Lin2014coco}, using mean Average Precision (mAP) at Intersection over Union (IoU) or Object Keypoint Similarity (OKS) thresholds. Specifically, we report mAP averaged over thresholds from 0.50 to 0.95 in increments of 0.05 (mAP@50:95).

\paragraph{Implementation details}
For each dataset the 2D-stacked and 3D variants are trained with identical schedules, augmentation, optimizer settings, and hardware, so that performance differences reflect architecture rather than training setup. The two experimental regimes do differ from each other: the real-world datasets are trained to convergence (100 epochs, $960{\times}960$), whereas the 60-run UCF101-24 ablation uses a shorter, lower-resolution budget (20 epochs, $320{\times}320$). Full optimizer hyperparameters, the augmentation pipeline, and hardware are given in~\ref{app:implementation}.

\subsection{Results}
We report model complexity (computational efficiency, including floating-point operations (FLOPs), parameter counts), baseline performance, and diagnostic ablations across both tasks. Together, these results reveal trade-offs between computational cost, raw accuracy, and robustness to temporal perturbations.
\paragraph{Computational Cost}
As shown in Fig.~\ref{fig:latency_tradeoff}, the nano and small variants do not exhibit a proportional change in inference latency, yet compute cost and parameter count rise sharply, especially at higher resolutions. We conduct all diagnostic experiments at nano scale for three reasons. First, the diagnostic framework is designed to compare \emph{architectural behaviours} — specifically, how temporal information flows through 2D stacked versus 3D convolutional pipelines — rather than to maximise absolute accuracy; nano scale isolates this architectural contrast with the fewest confounding capacity effects. Second, at nano scale the 3D and stacked-2D models have comparable parameter counts (Table~\ref{tab:core_baselines}), ensuring that observed differences in perturbation sensitivity reflect temporal integration strategy rather than raw model capacity. Third, both real-world datasets are moderately sized ($\sim5.5k$ and $\sim1.5k$ training samples), making larger model scales prone to overfitting and less representative of practical deployment scenarios in specialised domains. We note that the UCF101-24 ablation (Sec.~\ref{sec:ablation-ucf}) already demonstrates that the performance hierarchy among configurations is consistent across nano, small, and medium scales (Fig.~\ref{fig:ablation-bar}), supporting the generalisability of conclusions drawn at nano scale.

\begin{table}
\centering
\caption{\textbf{Baseline comparison of core model variants} mAP@50:95 (IoU or OKS), parameters, and GFLOPs for both tasks using full input sequences}
\label{tab:core_baselines}
\small
\setlength{\tabcolsep}{3.5pt}
\begin{tabular}{l cc c cc c}
\toprule
& \multicolumn{3}{c}{Operating Room} & \multicolumn{3}{c}{Dairy-Cow Pose} \\
\cmidrule(lr){2-4} \cmidrule(lr){5-7}
Model & mAP & Par.(M) & GF & OKS-mAP & Par.(M) & GF \\
\midrule
YOLOv8-2D (stacked) & .852 & 3.0 & 20.6 & .953 & 3.4 & 26.0 \\
YOLO-3D (ours) & .868 & 6.0 & 53.0 & .960 & 6.5 & 84.2 \\
\bottomrule
\end{tabular}
\end{table}

\begin{table}[t]
\centering
\caption{\textbf{Ablations under temporal perturbations} mAP@50:95 and delta ($\Delta$) from full-sequence baseline. Operating Room: IoU-based; Dairy-Cow: OKS-based}
\label{tab:ablation_merged}
\small
\setlength{\tabcolsep}{3pt}
\begin{tabular}{l cc cc cc cc}
\toprule
& \multicolumn{2}{c}{OR (2D)} & \multicolumn{2}{c}{OR (3D)} & \multicolumn{2}{c}{Cow (2D)} & \multicolumn{2}{c}{Cow (3D)} \\
\cmidrule(lr){2-3}\cmidrule(lr){4-5}\cmidrule(lr){6-7}\cmidrule(lr){8-9}
Perturbation & mAP & $\Delta$ & mAP & $\Delta$ & mAP & $\Delta$ & mAP & $\Delta$ \\
\midrule
None (full seq.) & .852 & — & .868 & — & .953 & — & .960 & — \\
FP-20\% & .859 & +.007 & .862 & $-$.006 & .900 & $-$.053 & .912 & $-$.048 \\
FP-50\% & .798 & $-$.054 & .845 & $-$.023 & .337 & $-$.616 & .540 & $-$.420 \\
FP-80\% & .796 & $-$.056 & .798 & $-$.070 & .182 & $-$.771 & .432 & $-$.528 \\
ES (even) & .765 & $-$.087 & .849 & $-$.019 & .320 & $-$.633 & .399 & $-$.561 \\
HL (hide last) & .088 & $-$.764 & .344 & $-$.524 & .246 & $-$.707 & .880 & $-$.080 \\
TS (full) & .781 & $-$.071 & .813 & $-$.055 & .483 & $-$.470 & .560 & $-$.400 \\
TS (exc.\ last) & .863 & +.011 & .864 & $-$.004 & .809 & $-$.144 & .669 & $-$.291 \\
FR-$k$ ($k{=}3$) & .831 & $-$.021 & .841 & $-$.027 & .954 & +.001 & .950 & $-$.010 \\
RD (60\%,$\times$8) & .867 & +.015 & .850 & $-$.018 & .943 & $-$.010 & .372 & $-$.588 \\
CL (last only) & .743 & $-$.109 & .767 & $-$.101 & .000 & $-$.953 & .390 & $-$.570 \\
\bottomrule
\end{tabular}
\end{table}

\paragraph{Baseline mean Average Precision}
Table~\ref{tab:core_baselines} shows that both stacked-2D and 3D models achieve high overall accuracy on both datasets. The 3D model provides a modest but consistent improvement in Operating Room mAP@50:95 (0.868 vs.\ 0.852) and Dairy-Cow pose OKS–mAP@50:95 (0.960 vs.\ 0.953), indicating better use of temporal cues for fine-grained localization. 

\paragraph{Diagnostic insights}
Table~\ref{tab:ablation_merged} reveals how the two architectures rely on temporal information in markedly different ways:

\textbf{Operating Room} Both models achieve near-maximal performance with full sequences. When early frames are occluded (FP-$p$), accuracy declines gradually for both. Under every-second occlusion (ES), however, the 2D model drops sharply—falling below its FP-80\% result despite having more visible frames—indicating a strong bias toward the latter part of the sequence. The 3D model remains closer to its FP-50\% performance, suggesting more distributed temporal use.  
Hide-Last (HL) reveals the clearest contrast: the 2D model nearly collapses (0.088), while the 3D model retains moderate accuracy (0.344), demonstrating its ability to recover information from earlier frames.  
Control (CL, last-only) and shuffle-except-last maintain high performance for both models, confirming that last-frame cues dominate this short 2-second setting. Full shuffle, frame replacement (FR), and resolution degradation (RD) have limited impact, implying that static appearance within this window is largely sufficient and temporal variation is minimal.

\textbf{Dairy-Cow Pose}
Temporal reasoning plays a much larger role here. Under increasing first-frame occlusion, both models degrade, but the 3D model remains more resilient at high occlusion (0.432 vs.\ 0.182 at FP-80\%).
Hide-Last (HL) makes the architectural gap starkest: the 2D model drops heavily (0.246), whereas the 3D model preserves high accuracy (0.880), reconstructing pose from temporal history. Full temporal shuffle (TS) reduces performance for both models, confirming sensitivity to temporal order. Shuffle-except-last again demonstrates the 2D model’s last-frame reliance, with accuracy recovering as long as the final frame remains intact. In the Control (CL, last-only) condition, the 2D model collapses entirely (0.0), while the 3D architecture maintains moderate accuracy (0.390), indicating reliance on broader temporal context.  
Finally, the 3D model is notably more sensitive to mid-sequence resolution degradation (RD-$q$), dropping to 0.372 compared to the 2D model’s near-invariance (0.943). This aligns with its reliance on motion continuity and mid-sequence detail, whereas the 2D model depends primarily on the last, uncorrupted frame.
\begin{figure}
    \centering

    \includegraphics[width=0.4\linewidth]{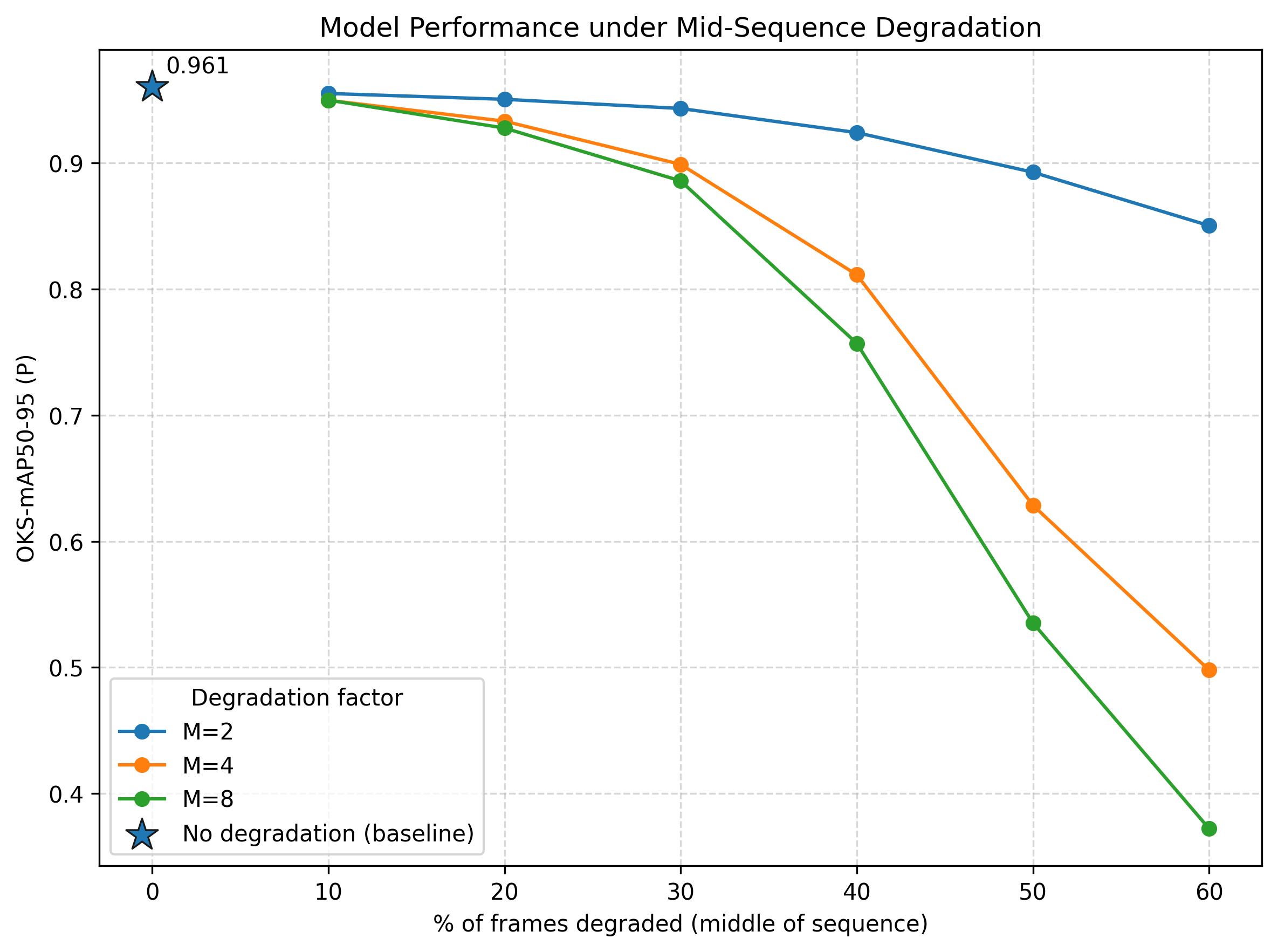}
    \includegraphics[width=0.4\linewidth]{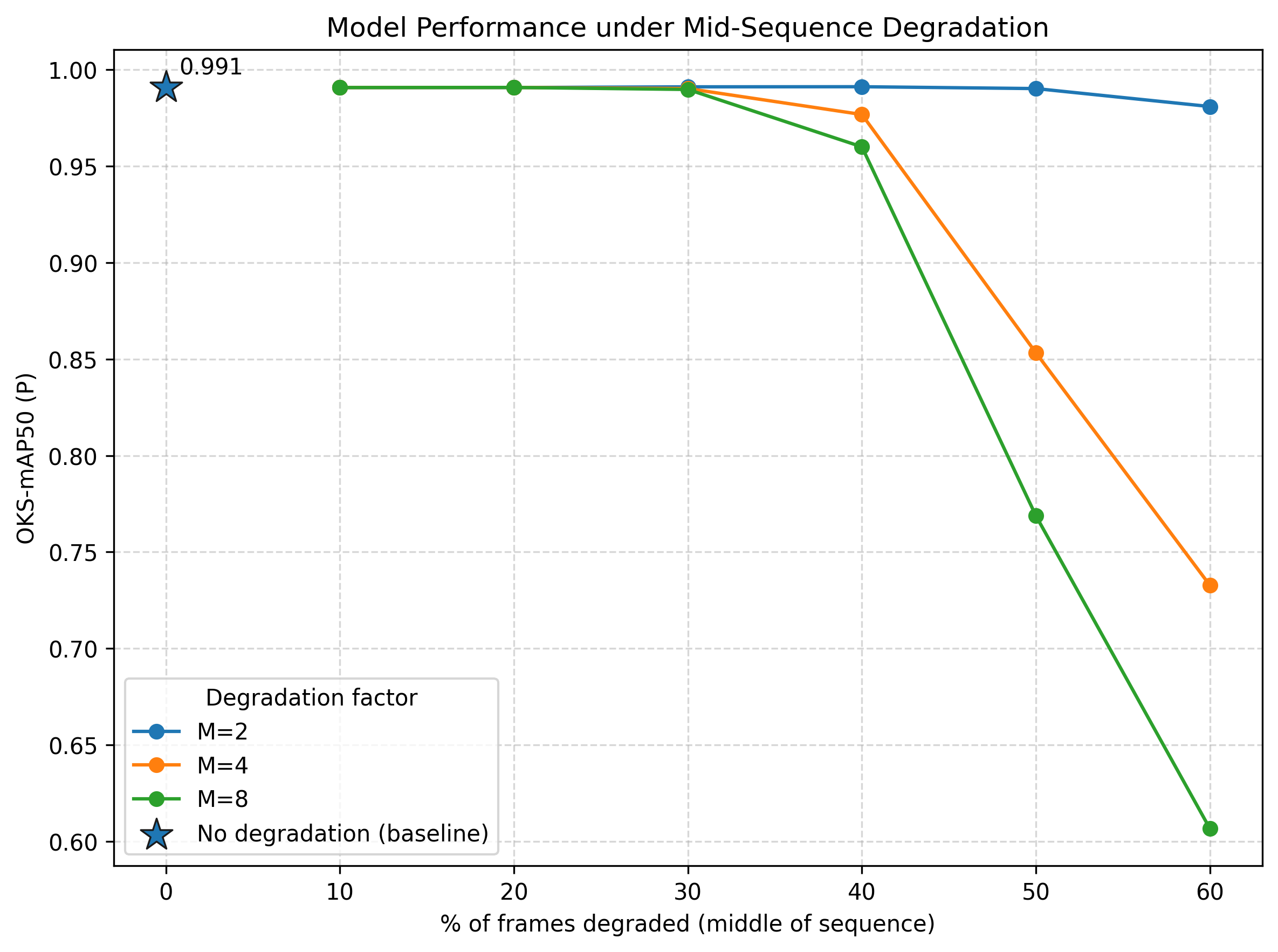}
    \caption{\textbf{Resolution Degradation Diagnosis (Dairy-Cow Pose)}
    We degrade the \emph{middle} portion of each sequence and report pose accuracy as a function of the percentage of degraded frames (x-axis) and degradation factor $M$ (curves; average-pool downsample by $M$ then nearest-neighbor upsample back). The star marks the no-degradation baseline. 
    \textbf{Left:} OKS–mAP@50:95. \textbf{Right:} OKS–mAP@50. 
    As both the \emph{duration} of the degraded segment and the \emph{severity} $M$ increase, performance drops monotonically, indicating strong reliance on fine mid-sequence detail (blocky/less-defined joints and limbs degrade skeleton quality). 
    Overall, the diagnosis exposes a fundamental trade-off: sequence-aware models benefit from temporal context but are more sensitive to widespread mid-sequence degradation}
    
    \label{fig:resolution-degradation}
\end{figure}

\textbf{Resolution Degradation Diagnosis (Dairy-Cow Pose)}
Figure \ref{fig:resolution-degradation} summarizes OKS-mAP@50:95 under mid-sequence resolution degradation. As degradation length and factor increase, performance drops monotonically for both models, but the 3D model collapses from 0.990 to 0.372 when the middle $60\%$ of frames are downsampled by ×8 and upsampled back. This is expected: the dairy-cow skeleton requires fine, mid-sequence detail (joints/limbs become blocky), and the 3D model suffers most when they are corrupted. In contrast, the stacked-2D model, which leans on the last frame, remains comparatively stable when that last frame is unaltered. Sequence-aware models benefit from temporal context but are more sensitive to widespread mid-sequence degradation; the diagnosis cleanly exposes this trade-off.

\textbf{Takeaway} The diagnostic framework reveals that the 2D stacked model succeeds when last-frame cues dominate, as in the Operating Room detection task, but offers little genuine temporal reasoning. The 3D backbone, while computationally heavier, demonstrates meaningful sequence integration: it maintains substantially higher accuracy even when the final frame is hidden, and it shows sensitivity to frame order and continuity, particularly in the dairy cow skeleton pose estimation task, where motion cues are essential. Still, the 3D model is not immune to degradation; its performance drops faster under heavy early frame occlusion and resolution reduction, highlighting its reliance on temporal continuity. Overall, the nano-scale 3D model strikes a practical balance, retaining stronger temporal modeling capacity than stacked 2D without the prohibitive cost of larger 3D variants.

\subsection{Architectural Ablation on UCF101-24}
\label{sec:ablation-ucf}

Our model contribution targets real-world challenges on domain-specific video datasets. By building on Ultralytics YOLOv8, we enable support for a variety of computer vision tasks and benefit from future ecosystem improvements. While our main focus remains on demonstrating performance in real-world medical and agricultural applications, we provide reproducible public benchmarks without prohibitive overhead by adopting {UCF101-24}~\cite{UCF101}: a subset containing 24 action classes from 3,207 videos. We utilize the dataset prepared by~\cite{Kopuklu2021YOWO}, which comprises ${\sim}338$k training clips and ${\sim}137$k test clips extracted from the annotated frames of the ${\sim}3.2$k videos in the dataset.

For reference, YOWO~\cite{Kopuklu2021YOWO} reports frame-mAP@50 of 0.610 (2D), 0.705 (3D), 0.730 (2D+3D), and 0.790 (2D+3D+CFAM), while YOWOv2~\cite{Zhiqiang2025YOWOv2} achieves 0.780 (nano), 0.805 (tiny), and 0.831 (medium) for 16-frame sequences, and 0.794 (nano), 0.830 (tiny), and 0.837 (medium) for 32-frame sequences.

We use UCF101-24 to conduct a systematic ablation study that isolates the contribution of each architectural component introduced in Sec.~\ref{sec:ModelArchitecture}. The ablation factorizes the design space into three axes: backbone type, neck fusion module, and temporal squeeze strategy, yielding 10 architectural configurations evaluated at three model scales.

\subsubsection{Ablation Design}
\label{sec:ablation-design}

Table~\ref{tab:ablation-configs} specifies the 10 configurations (A--J), each isolating one architectural change relative to a neighboring configuration. Crossed with three model scales (nano, small, medium), this yields 30 experiments. Each experiment is further evaluated at two sequence lengths (16 and 32 frames), producing 60 data points in total.

\begin{table*}[t]
\centering
\caption{\textbf{Ablation configurations} Each row differs from at least one neighbor in exactly one component.Configuration identifiers are non-contiguous (G and H are unused)}
\label{tab:ablation-configs}
\small
\setlength{\tabcolsep}{4pt}
\begin{tabular}{cllll}
\toprule
ID & Backbone & Neck Fusion & Squeeze & Isolates \\
\midrule
A & baseline & DynamicConcat & AvgPool & Reference \\
B & baseline & STAFLite & AvgPool & Attention fusion \\
C & baseline & STAFLite & ATF(learn) & Learned squeeze \\
D & temporal & STAFLite & ATF(learn) & Temporal backbone \\
D$^\dagger$ & temporal2 & STAFLite & ATF(learn) & Partial preservation \\
E & temporal & STAF (full) & ATF(learn) & Full vs.\ lite attention \\
E$^\dagger$ & temporal2 & STAF (full) & ATF(learn) & Full att.\ + partial backbone \\
F & temporal & DynamicConcat & AvgPool & Backbone alone \\
I & temporal & STAFLite & AvgPool & Simple squeeze \\
J & temporal & STAFLite & ATF(last) & Last-frame anchor \\
\bottomrule
\end{tabular}
\end{table*}

The three axes are:
\begin{itemize}
  \item \textbf{Backbone:} \emph{baseline} (temporal stride 2 at all five convolutions, Sec.~\ref{sec:backbone3d}), \emph{temporal} (temporal-preserving with stride $(1,2,2)$ at layers 3, 5, 7, Sec.~\ref{sec:temporal-backbone}), or \emph{temporal2} (partial preservation at layers 3 and 7 only, with layer 5 retaining stride $(2,2,2)$).
  \item \textbf{Neck fusion:} \emph{DynamicConcat} (Sec.~\ref{sec:backbone3d}), \emph{STAFLite} (lightweight gating variant, Sec.~\ref{sec:staf}), or \emph{STAF} (full factored spatiotemporal attention, Sec.~\ref{sec:staf}).
  \item \textbf{Temporal squeeze:} \emph{AvgPoolDepthSqueeze} (parameter-free adaptive average pooling to $D{=}1$, Sec.~\ref{sec:backbone3d}), \emph{ATF(\texttt{learn})} (learned soft attention over temporal axis, Sec.~\ref{sec:atf}), or \emph{ATF(\texttt{last})} (explicit last-frame anchor, Sec.~\ref{sec:atf}).
\end{itemize}


\subsubsection{Overall Performance}

Table~\ref{tab:ablation-medium} summarises the results for the medium-scale model, which provides the clearest separation between configurations.

\begin{table*}[t]
\centering
\caption{\textbf{Medium-scale ablation on UCF101-24} Best per column in \textbf{bold}. Values are mAP (\%). Abbreviations: base.\ = baseline backbone; temp.\ = temporal-preserving; temp.2 = partial temporal (layers 3, 7 only); DynCon.\ = DynamicConcat; STAFL.\ = STAFLite; AP = AvgPoolDepthSqueeze; ATF(l) = ATF(\texttt{learn}); ATF(la.) = ATF(\texttt{last}). Column headers denote sequence length and IoU threshold (e.g., 16f@.5 = 16 frames, mAP@50)}
\label{tab:ablation-medium}
\small
\setlength{\tabcolsep}{4pt}
\begin{tabular}{clll cccc}
\toprule
Cfg & Backbone & Fusion & Squeeze & 16f@.5 & 16f@.5:.95 & 32f@.5 & 32f@.5:.95 \\
\midrule
A & base. & DynCon. & AP & 79.8 & 49.4 & 81.1 & 49.2 \\
B & base. & STAFL. & AP & 80.3 & 49.7 & 78.9 & 48.1 \\
C & base. & STAFL. & ATF(l) & 79.5 & 49.1 & 81.2 & 49.1 \\
D & temp. & STAFL. & ATF(l) & 80.5 & 50.2 & \textbf{82.6} & 50.8 \\
D$^\dagger$ & temp.2 & STAFL. & ATF(l) & 79.5 & 49.3 & 80.9 & 49.5 \\
E & temp. & STAF & ATF(l) & 81.0 & 50.6 & 82.5 & 51.0 \\
E$^\dagger$ & temp.2 & STAF & ATF(l) & 80.0 & 49.5 & 80.3 & 49.4 \\
F & temp. & DynCon. & AP & \textbf{81.5} & \textbf{50.7} & \textbf{83.5} & \textbf{51.1} \\
I & temp. & STAFL. & AP & 78.0 & 48.9 & \textbf{82.8} & \textbf{51.2} \\
J & temp. & STAFL. & ATF(la.) & 79.8 & 49.7 & 81.8 & 50.0 \\
\bottomrule
\end{tabular}
\end{table*}

The top-performing configuration across both sequence lengths is \textbf{F} (temporal preserving backbone + DynamicConcat + AvgPool), reaching 83.5\% mAP@50 and 51.1\% mAP@50:95 at 32 frames. Configurations D and E, which add STAFLite, STAF fusion and ATF squeeze, are competitive at 82.5--82.6\% mAP@50 but do not surpass the simpler F configuration.

\begin{figure}
  \centering
  \includegraphics[width=0.85\textwidth]{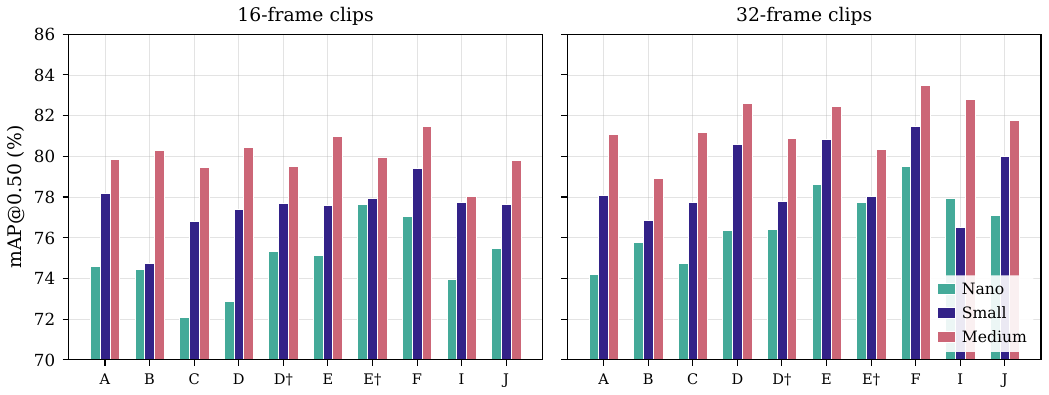}
  \caption{Comparison of mAP@50 (\%) across all 10 ablation configurations at three model scales (nano, small, medium) for 16-frame (left) and 32-frame (right) input clips on UCF101-24. Configurations A--C use the baseline temporal backbone; D, E, F, I, J use the temporal-preserving backbone; D$^\dagger$ and E$^\dagger$ use the partial temporal2 variant. Config F (temporal backbone + DynamicConcat + AvgPoolDepthSqueeze) achieves the highest mAP@50 at both sequence lengths}
  \label{fig:ablation-bar}
\end{figure}

Figure~\ref{fig:ablation-bar} presents the full results across all three model scales, revealing that the performance hierarchy remains consistent from nano through medium, with temporal preserving-backbone configurations (D, E, F, I, J) consistently outperforming excessive temporal downsampled baseline-backbone configurations (A, B, C), particularly at 32 frames.

\subsubsection{Component-wise Ablation}
\label{sec:component-ablation}

\begin{figure}
  \centering
  \includegraphics[width=0.7\linewidth]{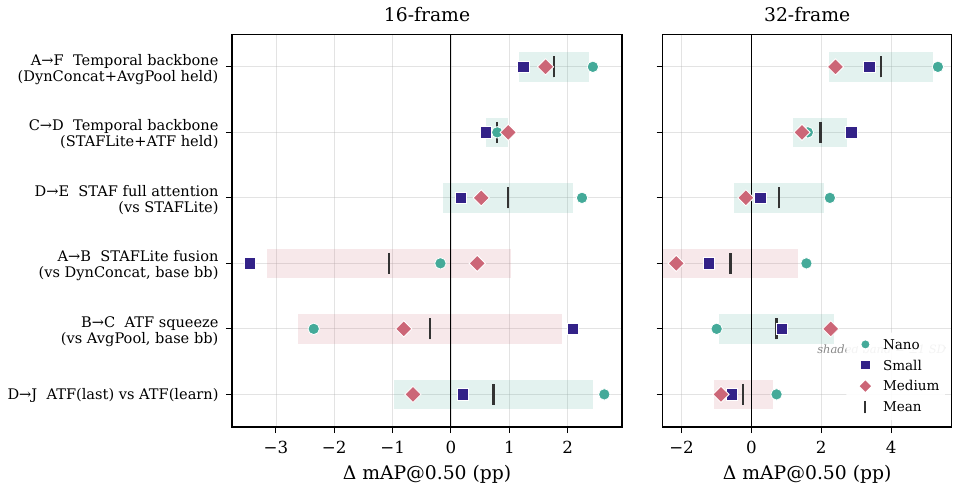}
  \caption{Component-wise contribution to mAP@50, measured as the difference (in percentage points) between configuration pairs that isolate a single architectural change. Values are averaged across nano, small, and medium scales. Left: 16-frame clips. Right: 32-frame clips. The temporal-preserving backbone provides the largest improvement in both settings, with the benefit doubling at 32 frames (+3.7\,pp vs +1.8\,pp). Light teal bars indicate improvements; rose bars indicate regressions}
  \label{fig:ablation-deltas}
\end{figure}

We isolate the contribution of each proposed module by comparing pairs of configurations that differ in exactly one component, averaging across the three model scales. Figure~\ref{fig:ablation-deltas} visualises these component-wise deltas.

\paragraph{Effect of the temporal-preserving backbone}
The temporal backbone is the single most impactful modification. Comparing A$\to$F (same DynamicConcat + AvgPool neck, only the backbone changes) yields $+1.8/+1.2$\,pp mAP@50/@50:95 at 16 frames and $+3.7/+2.6$\,pp at 32 frames. The benefit scales with sequence length, confirming the design hypothesis: under the baseline backbone's aggressive temporal striding ($\div32$), 16-frame clips reach the neck with $D{=}1$ at most pyramid levels, rendering temporal reasoning moot, whereas the temporal-preserving backbone (stride $(1,2,2)$ at layers 3, 5, 7) retains $D{=}4$ for 16-frame and $D{=}8$ for 32-frame input. Holding the full proposed neck constant (C$\to$D, STAFLite + ATF) gives smaller but consistent gains ($+0.8/+0.9$\,pp at 16f, $+2.0/+1.6$\,pp at 32f), indicating the backbone improvement is partly independent of the neck design.

\paragraph{Temporal2 vs.\ full temporal backbone.}
The temporal2 variant (preserving temporal depth at layers 3 and 7 only, while layer 5 retains stride $(2,2,2)$) consistently underperforms the full temporal backbone: at 32 frames and medium scale, D$^\dagger$ trails D by 1.7\,pp mAP@50 and E$^\dagger$ trails E by 2.2\,pp. Allowing temporal collapse at even one intermediate stage loses meaningful temporal context, confirming that consistent preservation across all deep layers is necessary.

\paragraph{Effect of attention-based fusion (STAFLite).}
Replacing DynamicConcat with STAFLite while holding other components constant yields mixed results: $-1.1$\,pp mAP@50 at 16f and $-0.6$\,pp at 32f on the baseline backbone (A$\to$B), and $-2.7/-2.4$\,pp on the temporal backbone (F$\to$I). STAFLite does not improve over simple DynamicConcat in either setting at this training budget (20 epochs), but serves as a prerequisite for the full STAF variant.

\paragraph{Effect of full spatio-temporal attention (STAF).}
Upgrading from STAFLite to the full STAF module (D$\to$E, temporal backbone) gives $+1.0/+0.5$\,pp mAP@50/@50:95 at 16 frames and $+0.8/+1.0$\,pp at 32 frames. The factored self-attention provides a small but consistent improvement over the lite variant, particularly for mAP@50:95 at 32 frames, suggesting it improves localisation quality when given richer temporal input.

\paragraph{Effect of adaptive temporal focus (ATF).}
Replacing AvgPool squeeze with ATF(\texttt{learn}) on the baseline backbone (B$\to$C) costs $-0.4/-0.6$\,pp at 16 frames but recovers $+0.7/+0.3$\,pp at 32 frames: ATF helps only where P3 retains $D{=}2$ since at 16 frames the other pyramid levels collapse to $D{=}1$ and the learned temporal weighting has no signal to operate on. Comparing the two ATF modes (D$\to$J, temporal backbone), learned soft attention and last-frame anchoring perform comparably---$+0.7$\,pp mAP@50 for ATF(\texttt{last}) at 16f, $-0.2$\,pp at 32f---with the anchor's slight edge on shorter clips possibly reflecting less benefit from a soft attention query.

\subsubsection{Effect of Sequence Length}
\begin{figure}
  \centering
  \includegraphics[width=0.55\linewidth]{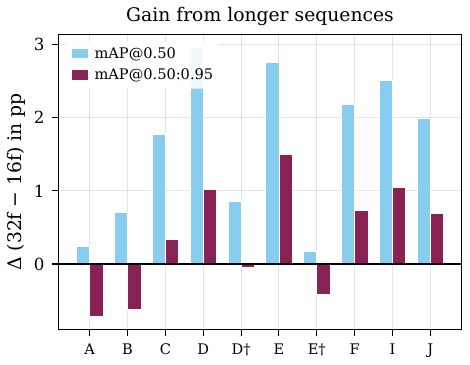}
  \caption{Average gain (in percentage points) from extending input clips
  from 16 to 32 frames, computed per configuration across all three model
  scales. Light blue: mAP@50; dark wine: mAP@50:95. Temporal-backbone
  configurations (D, E, F, I, J) benefit substantially from longer
  sequences ($+1.5$ to $+3.0$\,pp mAP@50), while baseline configurations
  (A, B) show minimal or negative change, confirming that the
  temporal-preserving backbone is prerequisite to exploiting longer clips}
  \label{fig:seq-length-gain}
\end{figure}

Longer input clips (32 vs.\ 16 frames) improve performance for nearly all
configurations, but the magnitude varies by architecture.
Figure~\ref{fig:seq-length-gain} quantifies this gain per configuration.
\begin{itemize}
  \item \textbf{Baseline backbone (A):} minimal gain ($+0.2$\,pp mAP@50
  averaged across scales), since temporal information is already discarded
  by the backbone.
  \item \textbf{Temporal backbone configs (D, E, F):} substantial gains of
  $+2.0$ to $+3.0$\,pp mAP@50, confirming that the temporal backbone is
  necessary to exploit longer sequences.
  \item \textbf{Configuration I} (temporal + STAFLite + AvgPool) shows the
  largest 32-frame benefit at $+2.5$\,pp averaged, with the medium scale
  alone gaining $+4.8$\,pp---suggesting the simple AvgPool squeeze becomes
  more effective when averaging over richer temporal features.
\end{itemize}
An experiment-level view of the same trend---temporal-backbone
configurations clustering above the $32\text{f}{=}16\text{f}$ diagonal,
most clearly at larger model scales---is provided in
Appendix~\ref{app:scatter} (Fig.~\ref{fig:scatter-16v32}).

\section{Discussion}
\label{sec:discussion}

Across both the real-world diagnostics and the UCF101-24 ablation, a single theme recurs: temporal modelling is valuable precisely, and only, where single-frame cues cannot resolve the underlying dynamics. The stacked-2D model reaches near-ceiling accuracy on the Operating Room task at far lower compute, confirming that last-frame appearance suffices in static or slow-changing scenes; the 3D model's advantage materialises in the motion-dependent dairy-cow setting, where it reconstructs predictions from earlier frames once the target frame is removed and the stacked-2D baseline collapses. Temporal convolutions are therefore not universally superior, and the diagnostic suite is what makes this distinction visible, it separates genuine sequence integration from incidental accuracy that raw mAP cannot tell apart.

The same diagnostics expose the cost of that integration. The 3D backbone's heightened sensitivity to mid-sequence resolution degradation (RD-$p$-$q$) is not an independent weakness but the direct corollary of its reliance on coherent motion cues: a model that genuinely reads across frames is necessarily more exposed when those frames are corrupted than one that leans on a single clean target frame. Read together, the robustness gains under occlusion and the losses under spatial degradation describe one and the same property from two directions. The tolerance
both architectures show to frame redundancy (FR-$k$) is consistent with this picture, with reduced temporal novelty acting as a mild regulariser rather than a meaningful perturbation in these short clips.

The ablation and the diagnostics play complementary roles rather than redundant ones. The ablation identifies \emph{what} architectural decision matters most---preserving non-degenerate temporal depth through the backbone, without which downstream fusion and focus modules operate
on collapsed single-frame features and the benefit of longer clips disappears. The diagnostics verify \emph{that} this decision translates into temporal reasoning rather than a accidental capacity effect. A notable consequence is that, under the training budget explored here, the simplest downstream configuration is sufficient once temporal information is preserved upstream: the bottleneck is information \emph{availability} in the backbone, not the sophistication of the temporal processing that
follows. The learned attention modules still contribute consistent, if smaller, localisation gains when given a sufficient temporal signal, leaving open whether longer schedules would shift the balance further in their favour---a question we return to in Sec.~\ref{sec:limitations}.

These observations carry two practical implications. As an evaluation instrument, the perturbation suite exposes robustness gaps that raw accuracy conceals when single-frame cues dominate, making it suited to deployment-critical model assessment. As a design guide, it reframes the choice between single-image, stacked-2D, and 3D architectures as one to be settled by diagnostic evidence against task requirements, weighing temporal robustness against compute and latency, rather than by mAP alone.

\subsection{Limitations and Future Directions}
\label{sec:limitations}

While our 3D variant demonstrates genuine temporal reasoning, it introduces notable computational overhead, especially at larger scales and longer sequence lengths. Future work should explore optimizing the backbone (e.g., a lighter alternative to the C2f3d layer) further while preserving temporal information. 
Additionally, systematic studies on varying sequence length and sampling rates would further clarify how temporal integration scales across applications. Broader evaluations on diverse public benchmarks would further test generalizability; e.g., TITAN-Human Action benchmark \cite{ZHONG2025TitanBenchmark}, which targets multi-person spatial–temporal action detection in urban driving scenes where temporal cues are critical, is a natural next target for applying both YOLO-3D and the TemporalLens diagnostics. Extensions to volumetric modalities such as MRI could also highlight new application domains and strengthen claims of generalizability.

Our diagnostic framework is designed to be model-agnostic in the sense that it applies to any detector that receives a single temporal input tensor and produces predictions from it. This covers the dominant design pattern in single-stage video detection, including the stacked-frame 2D approach proposed by Corsel~et~al.~\cite{Corsel_2023_WACV}, van Leeuwen~et~al.~\cite{vanLeeuwen2024TemporalYOLOv8} and Van~Lier~et~al.~\cite{van_Lier_2025_WACV}, which serves as our stacked 2D baseline. However, dual-branch architectures such as YOWO~\cite{Kopuklu2021YOWO, Zhiqiang2025YOWOv2} fall outside this scope: their 2D branch receives the target frame directly as a separate input, so perturbations applied to the 3D branch (e.g., Hide-Last, Control-Last) leave the 2D pathway—and its access to the target frame—intact, rendering the perturbation semantics ill-defined without further architectural adaptation. Extending the diagnostic protocol to such multi-input designs, for instance, by jointly perturbing both branches or by selectively masking the 2D target-frame input, is a promising direction for future work.

\textbf{Training compute and schedule asymmetry} The two experimental regimes use deliberately different budgets. The UCF101-24 ablation spans 60 training runs (10 configurations × 3 scales × 2 clip lengths), which makes long schedules prohibitively expensive, so we cap it at 20 epochs, sufficient to establish the relative ordering of configurations that our analysis targets. The two smaller real-world datasets are inexpensive enough to train to convergence (100 epochs), where even at the nano scale and under the stricter mAP@50:95 criterion, both models reach high accuracy (Table~\ref{tab:core_baselines}), indicating the comparison there is not training-limited. Because the diagnostic suite probes temporal integration rather than augmentation strength or absolute accuracy, this asymmetry does not confound the architectural contrast it is designed to measure. The 20-epoch cap does, however, leave open whether longer schedules would let the learned fusion and focus modules (STAF, ATF) overtake the simpler configuration F. Compute-efficient training is a natural enabler: the anti-forgetting sampling of Xie et al.~\cite{Xie_2026_CVPR_YOLO_Efficient_Training}, for instance, reports more than 1.4× YOLO training speedups with no loss in accuracy by skipping already-learned images each epoch. Such methods would make both longer ablation schedules and the extension of YOLO-3D to larger, higher-resolution datasets, where the cost of 3D convolutions is most acute, substantially more tractable.

\section{Conclusion}
\label{sec:conclusion}

We presented YOLO-3D, a family of modular spatiotemporal detectors built on YOLOv8 for real-time video detection, and a model-agnostic diagnostic framework that probes whether video detectors genuinely reason over time or collapse to single-frame cues.

A systematic ablation on UCF101-24 (60 experiments spanning 10 configurations, 3 model scales, and 2 sequence lengths) established a clear hierarchy of design importance: the temporal-preserving backbone, which retains non-degenerate temporal depth at mid-to-late feature pyramid levels, is the single most impactful component, yielding up to $+3.7$\,pp mAP@50 and scaling its benefit with clip length. Notably, the simplest downstream modules (concatenation-based fusion and average pooling) suffice to achieve the best overall accuracy when temporal information is properly preserved in the backbone, suggesting that \emph{preserving} temporal signal matters more than sophisticated temporal \emph{processing}, at least on the chosen dataset and the training budgets explored here.

The diagnostic framework, applied across surgical scene analysis and articulated animal pose estimation, confirmed that these architectural gains correspond to genuine temporal integration: 3D models maintain predictive capability when the target frame is removed and exhibit sensitivity to frame order and mid-sequence degradation, whereas stacked-frame 2D models collapse under the same conditions. These behavioural differences are invisible to standard mAP evaluation, underscoring the practical value of controlled perturbation-based diagnostics for model selection and deployment.

Together, the architectural study and the diagnostic protocol provide complementary tools for the design and evaluation of temporally aware single-stage detectors.

\section*{Acknowledgements}
\noindent This research was supported by the Federal Ministry of Research, Technology and Space (BMFTR) through projects K3VR [grant: 13N16388] and Kiara [grant: 16SV9036]. Thanks to Daniia Vergazova for her assistance with software development.

\section*{Data Availability}
The UCF101-24 benchmark dataset used in the architectural ablation study is publicly available through its original release. The two domain-specific datasets — surgical scenes from kidney transplantation and dairy-cow pose estimation from veterinary farm surveillance — contain confidential data collected under institutional agreements and cannot be made publicly available due to privacy and contractual restrictions. Requests for access to these datasets may be directed to the corresponding author and will be considered subject to the approval of the data-providing institutions. The full source code for the YOLO-3D model family and the diagnostic perturbation framework (TemporalLens) is publicly released at our \href{https://gitlab.hhi.fraunhofer.de/dawoud/temporal-lens}{GitLab repository} to enable the research community to reproduce the methodology and apply it to their own datasets.

\section*{CRediT authorship contribution statement}
\textbf{Karam Tomotaki-Dawoud:} Conceptualization of this study, Methodology, Writing - Original Draft.\textbf{Anna
Hilsmann:} Writing - Review Editing. \textbf{Peter Eisert:} Writing - Review Editing. \textbf{Sebastian Bosse:} Supervision, Writing - Review Editing

\section*{Declaration of generative AI and AI-assisted technologies in the manuscript preparation process}
\noindent During the preparation of this work the authors used Claude (Anthropic) to assist with editing, language clarity and grammar improvement of this manuscript.

\section*{Declaration of competing interest}
\noindent The authors declare that they have no known competing financial interests or personal relationships that could have appeared to influence the work reported in this paper.

\bibliographystyle{elsarticle-num} 
\bibliography{manuscript_ref}

@string(CVPR= {IEEE Conf. Comput. Vis. Pattern Recog.})

@string(ICCV= {Int. Conf. Comput. Vis.})

@string(ECCV= {Eur. Conf. Comput. Vis.})

@string(NIPS= {Adv. Neural Inform. Process. Syst.})

@string(ACCV  = {ACCV})

@string(AAAI = {AAAI})

@string(CVPRW= {IEEE Conf. Comput. Vis. Pattern Recog. Worksh.})

@string(CVPR  = {CVPR})

@string(ICCV  = {ICCV})

@string(ECCV  = {ECCV})

@string(NIPS  = {NeurIPS})

@string(CVPRW= {CVPRW})

@inproceedings{Zhiqiang2025YOWOv2,
	title        = {YOWOv2: A Stronger yet Efficient Multi-level Detection Framework for Real-Time Spatio-Temporal Action Detection},
	author       = {Jiang, Zhiqiang and Yang, Jianhua and Jiang, Nan and Liu, Shuaiyan and Xie, Tao and Zhao, Lijun and Li, Ruifeng},
	year         = 2025,
	booktitle    = {Intelligent Robotics and Applications},
	publisher    = {Springer Nature Singapore},
	address      = {Singapore},
	editor       = {Lan, Xuguang and Mei, Xuesong and Jiang, Caigui and Zhao, Fei and Tian, Zhiqiang}
}

@article{UCF101,
	title        = {{UCF101:} {A} Dataset of 101 Human Actions Classes From Videos in The Wild},
	author       = {Khurram Soomro and Amir Roshan Zamir and Mubarak Shah},
	year         = 2012,
	journal      = {CoRR},
	volume       = {abs/1212.0402},
	eprinttype   = {arXiv}
}

@software{yolov8_ultralytics,
	title        = {{Ultralytics YOLO}},
	author       = {Jocher, Glenn and Qiu, Jing and Chaurasia, Ayush},
	year         = 2023,
	url          = {https://github.com/ultralytics/ultralytics},
	license      = {AGPL-3.0},
	version      = {8.2.26}
}

@inproceedings{Tran2015C3D,
	title        = {Learning Spatiotemporal Features with 3D Convolutional Networks},
	author       = {Du Tran and Lubomir D. Bourdev and Rob Fergus and Lorenzo Torresani and Manohar Paluri},
	year         = 2015,
	booktitle    = {ICCV}
}

@inproceedings{Carreira2017I3D,
	title        = {Quo Vadis, Action Recognition? A New Model and the Kinetics Dataset},
	author       = {Joao Carreira and Andrew Zisserman},
	year         = 2017,
	booktitle    = {CVPR}
}

@inproceedings{Hara2018Res3D,
	title        = {Can Spatiotemporal 3D CNNs Retrace the History of 2D CNNs and ImageNet?},
	author       = {Kensho Hara and Hirokatsu Kataoka and Yutaka Satoh},
	year         = 2018,
	booktitle    = {CVPR}
}

@inproceedings{Tran2018Closer,
	title        = {A Closer Look at Spatiotemporal Convolutions for Action Recognition},
	author       = {Du Tran and Heng Wang and Lorenzo Torresani and Jamie Ray and Yann LeCun and Manohar Paluri},
	year         = 2018,
	booktitle    = {CVPR}
}

@inproceedings{Feichtenhofer2019SlowFast,
	title        = {SlowFast Networks for Video Recognition},
	author       = {Christoph Feichtenhofer and Haoqi Fan and Jitendra Malik and Kaiming He},
	year         = 2019,
	booktitle    = {ICCV}
}

@inproceedings{Feichtenhofer2020X3D,
	title        = {{X3D}: Expanding Architectures for Efficient Video Recognition},
	author       = {Christoph Feichtenhofer},
	year         = 2020,
	booktitle    = {CVPR}
}

@article{Kopuklu2021YOWO,
	title        = {You Only Watch Once: A Unified CNN Architecture for Real-Time Spatiotemporal Action Localization},
	author       = {Okan K\"{o}p\"{u}kl\"{u} and Xiangyu Wei and Gerhard Rigoll},
	year         = 2021,
	journal      = {arXiv preprint arXiv:1911.06644},
	note         = {v5 Oct 2021}
}

@article{Redmon2018YOLOv3,
	title        = {{YOLOv3}: An Incremental Improvement},
	author       = {Joseph Redmon and Ali Farhadi},
	year         = 2018,
	journal      = {arXiv preprint arXiv:1804.02767}
}

@article{Bochkovskiy2020YOLOv4,
	title        = {{YOLOv4}: Optimal Speed and Accuracy of Object Detection},
	author       = {Alexey Bochkovskiy and Chien-Yao Wang and Hong-Yuan Mark Liao},
	year         = 2020,
	journal      = {arXiv preprint arXiv:2004.10934}
}

@inproceedings{Zeiler_occlusion,
	title        = {Visualizing and Understanding Convolutional Networks},
	author       = {Zeiler, Matthew D. and Fergus, Rob},
	year         = 2014,
	booktitle    = {ECCV},
	editor       = {Fleet, David and Pajdla, Tomas and Schiele, Bernt and Tuytelaars, Tinne}
}

@inproceedings{NEURIPS2019_Hooker,
	title        = {A Benchmark for Interpretability Methods in Deep Neural Networks},
	author       = {Hooker, Sara and Erhan, Dumitru and Kindermans, Pieter-Jan and Kim, Been},
	year         = 2019,
	booktitle    = {NeurIPS},
	volume       = 32,
	editor       = {H. Wallach and H. Larochelle and A. Beygelzimer and F. d\textquotesingle Alch\'{e}-Buc and E. Fox and R. Garnett}
}

@inproceedings{Agarwal_Occ2020,
	title        = {Explaining Image Classifiers by Removing Input Features Using Generative Models},
	author       = {Agarwal, Chirag and Nguyen, Anh},
	year         = 2020,
	booktitle    = {ACCV},
	editor       = {Ishikawa, Hiroshi and Liu, Cheng-Lin and Pajdla, Tomas and Shi, Jianbo}
}

@article{Samek2021,
	title        = {Explaining Deep Neural Networks and Beyond: A Review of Methods and Applications},
	author       = {Samek, Wojciech and Montavon, Grégoire and Lapuschkin, Sebastian and Anders, Christopher J. and Müller, Klaus-Robert},
	year         = 2021,
	journal      = {Proceedings of the IEEE},
	volume       = 109
}

@inproceedings{dawoud2023,
	title        = {Human-Centered Evaluation of XAI Methods},
	author       = {Dawoud, Karam and Samek, Wojciech and Eisert, Peter and Lapuschkin, Sebastian and Bosse, Sebastian},
	year         = 2023,
	booktitle    = {ICDMW},
	volume       = {},
	number       = {}
}

@inproceedings{Lin2017FPN,
	title        = {Feature Pyramid Networks for Object Detection},
	author       = {Lin, Tsung-Yi and Dollár, Piotr and Girshick, Ross and He, Kaiming and Hariharan, Bharath and Belongie, Serge},
	year         = 2017,
	booktitle    = {CVPR},
	volume       = {},
	number       = {}
}

@inproceedings{Liu2018PANet,
	title        = {Path Aggregation Network for Instance Segmentation},
	author       = {Shu Liu and Lu Qi and Haifang Qin and Jianping Shi and Jiaya Jia},
	year         = 2018,
	booktitle    = {CVPR}
}

@inproceedings{Bertasius2021TimeSformer,
	title        = {Is Space-Time Attention All You Need for Video Understanding?},
	author       = {Gedas Bertasius and Heng Wang and Lorenzo Torresani},
	year         = 2021,
	booktitle    = {ICML}
}

@inproceedings{Arnab2021ViViT,
	title        = {ViViT: A Video Vision Transformer},
	author       = {Arnab, Anurag and Dehghani, Mostafa and Heigold, Georg and Sun, Chen and Lučić, Mario and Schmid, Cordelia},
	year         = 2021,
	booktitle    = {ICCV},
	volume       = {},
	number       = {}
}

@inproceedings{Liu2022VideoSwin,
	title        = {Video Swin Transformer},
	author       = {Liu, Ze and Ning, Jia and Cao, Yue and Wei, Yixuan and Zhang, Zheng and Lin, Stephen and Hu, Han},
	year         = 2022,
	booktitle    = {CVPR},
	volume       = {},
	number       = {}
}

@article{Shi2023YOLOV,
	title        = {YOLOV: Making Still Image Object Detectors Great at Video Object Detection},
	author       = {Shi, Yuheng and Wang, Naiyan and Guo, Xiaojie},
	year         = 2023,
	journal      = {Proceedings of the AAAI Conference on Artificial Intelligence},
	volume       = 37,
	number       = 2
}

@article{vanLeeuwen2024TemporalYOLOv8,
	title        = {Toward Versatile Small Object Detection with Temporal-YOLOv8},
	author       = {van Leeuwen, Martin C. and Fokkinga, Ella P. and Huizinga, Wyke and Baan, Jan and Heslinga, Friso G.},
	year         = 2024,
	journal      = {Sensors},
	volume       = 24,
	number       = 22,
	article-number = 7387,
	pubmedid     = 39599163
}

@inproceedings{Tong2022VideoMAE,
	title        = {VideoMAE: masked autoencoders are data-efficient learners for self-supervised video pre-training},
	author       = {Tong, Zhan and Song, Yibing and Wang, Jue and Wang, Limin},
	year         = 2022,
	booktitle    = {NeurIPS},
	series       = {NIPS '22},
	articleno    = 732
}

@inproceedings{Wang2020CSPNet,
	title        = {CSPNet: A New Backbone that can Enhance Learning Capability of CNN},
	author       = {Wang, Chien-Yao and Mark Liao, Hong-Yuan and Wu, Yueh-Hua and Chen, Ping-Yang and Hsieh, Jun-Wei and Yeh, I-Hau},
	year         = 2020,
	booktitle    = {CVPRW},
	volume       = {},
	number       = {}
}

@inproceedings{Suzuki_2018_ECCVWS,
	title        = {Learning Spatiotemporal 3D Convolution with Video Order Self-supervision},
	author       = {Suzuki, Tomoyuki and Itazuri, Takahiro and Hara, Kensho and Kataoka, Hirokatsu},
	year         = 2019,
	booktitle    = {ECCVW},
	editor       = {Leal-Taix{\'e}, Laura and Roth, Stefan}
}

@inproceedings{Lin2014coco,
	title        = {Microsoft COCO: Common Objects in Context},
	author       = {Lin, Tsung-Yi and Maire, Michael and Belongie, Serge and Hays, James and Perona, Pietro and Ramanan, Deva and Doll{\'a}r, Piotr and Zitnick, C. Lawrence},
	year         = 2014,
	booktitle    = {ECCV},
	editor       = {Fleet, David and Pajdla, Tomas and Schiele, Bernt and Tuytelaars, Tinne}
}

@inproceedings{Goyal2017something,
	title        = {The “Something Something” Video Database for Learning and Evaluating Visual Common Sense},
	author       = {Goyal, Raghav and Kahou, Samira Ebrahimi and Michalski, Vincent and Materzynska, Joanna and Westphal, Susanne and Kim, Heuna and Haenel, Valentin and Fruend, Ingo and Yianilos, Peter and Mueller-Freitag, Moritz and Hoppe, Florian and Thurau, Christian and Bax, Ingo and Memisevic, Roland},
	year         = 2017,
	booktitle    = {ICCV},
	volume       = {},
	number       = {}
}

@inproceedings{van_Lier_2025_WACV,
	title        = {{ Evaluation of Spatio-Temporal Small Object Detection in Real-World Adverse Weather Conditions }},
	author       = {Van Lier, Michel and Van Leeuwen, Martin and Van Manen, Bastian and Kampmeijer, Leo and Boehrer, Nicolas},
	year         = 2025,
	booktitle    = {WACVW}
}

@inproceedings{Corsel_2023_WACV,
	title        = {Exploiting Temporal Context for Tiny Object Detection},
	author       = {Corsel, Christof W. and van Lier, Michel and Kampmeijer, Leo and Boehrer, Nicolas and Bakker, Erwin M.},
	year         = 2023,
	booktitle    = {WACVW},
	volume       = {},
	number       = {}
}

@inproceedings{Wang23VideoMAEV2,
	title        = {VideoMAE V2: Scaling Video Masked Autoencoders with Dual Masking},
	author       = {Wang, Limin and Huang, Bingkun and Zhao, Zhiyu and Tong, Zhan and He, Yinan and Wang, Yi and Wang, Yali and Qiao, Yu},
	year         = 2023,
	booktitle    = {CVPR},
	volume       = {},
	number       = {}
}

@inproceedings{Hashmi_2025_FAIM,
	title        = {Beyond Boxes: Mask-Guided Spatio-Temporal Feature Aggregation for Video Object Detection},
	author       = {Hashmi, Khurram Azeem and Sheikh, Talha Uddin and Stricker, Didier and Afzal, Muhammad Zeshan},
	year         = 2025,
	booktitle    = {WACV}
}

@inproceedings{Sarkar_MaskVD2025,
	title        = {MaskVD: Region Masking for Efficient Video Object Detection},
	author       = {Sarkar, Sreetama and Datta, Gourav and Kundu, Souvik and Zheng, Kai and Bhattacharyya, Chirayata and Beerel, Peter A.},
	year         = 2025,
	booktitle    = {WACV},
	volume       = {},
	number       = {}
}

@inproceedings{Tomoki_AOSA23,
	title        = {Visually explaining 3D-CNN predictions for video classification with an adaptive occlusion sensitivity analysis},
	author       = {Uchiyama, Tomoki and Sogi, Naoya and Niinuma, Koichiro and Fukui, Kazuhiro},
	year         = 2023,
	booktitle    = {WACV},
	volume       = {},
	number       = {}
}

@inproceedings{katharopoulos2020transformers,
	title        = {Transformers are rnns: Fast autoregressive transformers with linear attention},
	author       = {Katharopoulos, Angelos and Vyas, Apoorv and Pappas, Nikolaos and Fleuret, Fran{\c{c}}ois},
	year         = 2020,
	booktitle    = {International conference on machine learning},
	organization = {PMLR}
}

@inproceedings{zheng2023efficient,
	title        = {Efficient Attention via Control Variates},
	author       = {Lin Zheng and Jianbo Yuan and Chong Wang and Lingpeng Kong},
	year         = 2023,
	booktitle    = {International Conference on Learning Representations}
}

@inproceedings{zheng2022linear,
	title        = {Linear complexity randomized self-attention mechanism},
	author       = {Lin Zheng and Chong Wang and Lingpeng Kong},
	year         = 2022,
	booktitle    = {International Conference on Machine Learning},
	organization = {PMLR}
}

@inproceedings{Hu2018SENet,
	title        = {Squeeze-and-Excitation Networks},
	author       = {Hu, Jie and Shen, Li and Sun, Gang},
	year         = 2018,
	booktitle    = {2018 IEEE/CVF Conference on Computer Vision and Pattern Recognition},
	volume       = {},
	number       = {}
}

@inproceedings{Chen_2021_CVPR,
	title        = {Deep Analysis of CNN-Based Spatio-Temporal Representations for Action Recognition},
	author       = {Chen, Chun-Fu Richard and Panda, Rameswar and Ramakrishnan, Kandan and Feris, Rogerio and Cohn, John and Oliva, Aude and Fan, Quanfu},
	year         = 2021,
	month        = {June},
	booktitle    = {Proceedings of the IEEE/CVF Conference on Computer Vision and Pattern Recognition (CVPR)}
}

@inproceedings{Liu_2021_ICCV,
	title        = {TAM: Temporal Adaptive Module for Video Recognition},
	author       = {Liu, Zhaoyang and Wang, Limin and Wu, Wayne and Qian, Chen and Lu, Tong},
	year         = 2021,
	month        = {October},
	booktitle    = {Proceedings of the IEEE/CVF International Conference on Computer Vision (ICCV)}
}

@inproceedings{Qing_2023_ICCV,
	title        = {Disentangling Spatial and Temporal Learning for Efficient Image-to-Video Transfer Learning},
	author       = {Qing, Zhiwu and Zhang, Shiwei and Huang, Ziyuan and Zhang, Yingya and Gao, Changxin and Zhao, Deli and Sang, Nong},
	year         = 2023,
	month        = {October},
	booktitle    = {Proceedings of the IEEE/CVF International Conference on Computer Vision (ICCV)}
}

@inproceedings{Xie_2026_CVPR_YOLO_Efficient_Training,
	title        = {Does YOLO Really Need to See Every Training Image in Every Epoch?},
	author       = {Xie, Xingxing and Dong, Jiahua and Han, Junwei and Cheng, Gong},
	year         = 2026,
	month        = {June},
	booktitle    = {Proceedings of the IEEE/CVF Conference on Computer Vision and Pattern Recognition (CVPR)}
}

@article{ZHANG2020LearningMotion,
	title        = {Learning motion representation for real-time spatio-temporal action localization},
	author       = {Dejun Zhang and Linchao He and Zhigang Tu and Shifu Zhang and Fei Han and Boxiong Yang},
	year         = 2020,
	journal      = {Pattern Recognition},
	volume       = 103
}

@article{LIU2021ACDnet,
	title        = {ACDnet: An action detection network for real-time edge computing based on flow-guided feature approximation and memory aggregation},
	author       = {Yu Liu and Fan Yang and Dominique Ginhac},
	year         = 2021,
	journal      = {Pattern Recognition Letters},
	volume       = 145
}

@article{ZHONG2025TitanBenchmark,
	title        = {A benchmark dataset and semantics-guided detection network for spatial–temporal human actions in urban driving scenes},
	author       = {Fujin Zhong and Yini Wu and Hong Yu and Guoyin Wang and Zhantao Lu},
	year         = 2025,
	journal      = {Pattern Recognition},
	volume       = 158
}

\newpage

\appendix

\section{Temporal-Preserving Backbone: Cost Analysis}
\label{app:cost}

The temporal-preservation strategy involves a favourable trade-off across three distinct cost dimensions (Table~\ref{tab:cost-tradeoff}).

First, switching from $3{\times}3{\times}3$ to $1{\times}3{\times}3$ kernels at three backbone layers reduces the per-filter parameter count by $67\%$ at each affected layer, yielding an $18\%$ reduction in backbone parameters. This translates to a $14.1\%$ reduction in total
model parameters when the neck is otherwise unchanged (config F), and a $6.4\%$ reduction once the new fusion and squeeze modules are added, since those do not fully offset the kernel savings.

Second, because feature tensors retain $D{>}1$ through the neck, activation memory per sample increases by $30$--$46\%$ depending on clip length. This is the primary cost and may require modest batch-size adjustments on memory-constrained hardware.

Third, forward-pass latency is essentially unchanged at $T{=}8$, where the extra temporal dimension is small, and increases at longer clips, scaling with the additional volume processed by the C2f3d bottleneck blocks.

Crucially, the activation-memory overhead is constant in relative terms across model scales ($n$, $s$, $m$), making the trade-off predictable when scaling the architecture.

\begin{table}[h]
\centering
\caption{Cost comparison between baseline and temporal-preserving backbone (configs A and F from Sec.~\ref{sec:ablation-ucf}, n-scale, $320{\times}320$, 24-class head). The temporal-preserving variant reduces parameters while increasing activation memory due to retained temporal depth}
\label{tab:cost-tradeoff}
\small
\setlength{\tabcolsep}{4pt}
\begin{tabular}{lccc}
\toprule
Metric & Baseline (A) & Temporal config F($\Delta$) & Temporal config D($\Delta$) \\
\midrule
Total parameters & 8.71\,M & 7.5\,M ($-14.1\%$) & 8.20\,M ($-5.9\%$) \\
\midrule
Activation memory ($T{=}8$) & 16.2\,MB & 21.1\,MB ($+30\%$)  & 21.1\,MB ($+30\%$) \\
Activation memory ($T{=}16$) & 28.5\,MB & 40.8\,MB ($+43\%$) & 40.8\,MB ($+43\%$)\\
Activation memory ($T{=}32$) & 54.9\,MB & 80.1\,MB ($+46\%$)&  80.1\,MB ($+46\%$) \\
\bottomrule
\end{tabular}
\end{table}

\section{TemporalLens Perturbation Specifications}
\label{app:perturbations}

This appendix gives the formal masking operators for the seven
perturbations summarized in Table~\ref{tab:perturbations}, together with
their per-probe diagnostic expectations and the generalization to an
arbitrary target-frame index. Throughout, $\mathbf{x} \in
\mathbb{R}^{C \times T \times H \times W}$ denotes the input clip and
$\boldsymbol{\mu}$ the dataset-wide RGB mean frame.

\paragraph{(1) First-$p$-Fraction (FP-$p$)}
\textit{Definition:} replace the first $\lceil pT\rceil$ frames with
$\boldsymbol{\mu}$: $\mathbf{x}[:,\, 0:\lceil pT\rceil] = \boldsymbol{\mu}$,
with $p \in \{0.20, 0.50, 0.80\}$.
\textit{Expectation:} if early temporal context contributes
meaningfully, performance should decline with increasing $p$, though the
trajectory depends on where the occlusion boundary falls relative to the
labeled target frame.

\paragraph{(2) Every-Second Occlusion (ES-$s$)}
\textit{Definition:} replace alternating frames with $\boldsymbol{\mu}$,
starting at offset $s \in \{0, 1\}$: $\mathbf{x}[:,\, s::2] =
\boldsymbol{\mu}$.
\textit{Expectation:} disrupting temporal continuity should hurt models
that integrate motion. Unlike contiguously occluding the first 50\% of
frames, this interleaved pattern reduces redundancy and may yield
comparatively better performance.

\paragraph{(3) Hide-Last (HL)}
\textit{Definition:} replace only the final frame with
$\boldsymbol{\mu}$: $\mathbf{x}[:,\, T{-}1] = \boldsymbol{\mu}$.
\textit{Expectation:} severe degradation signals heavy last-frame
reliance; 3D convolutional models may retain partial accuracy by
leveraging earlier frames.

\paragraph{(4) Control: Last-Only (CL)}
\textit{Definition:} replace all frames except the last with
$\boldsymbol{\mu}$: $\mathbf{x}[:,\, :T{-}1] = \boldsymbol{\mu}$.
\textit{Expectation:} establishes an upper bound on performance
achievable without temporal information, serving as a baseline for
quantifying last-frame dependence.

\paragraph{(5) Temporal Shuffle (TS / TS-Except-Last)}
\textit{Definition:} randomly permute all frames using a fixed seed:
$\mathbf{x}[:,\, t] \leftarrow \mathbf{x}[:,\, \pi(t)]$. In the
\emph{TS-Except-Last} variant the final frame remains in its original
position.
\textit{Expectation:} full shuffling should strongly degrade models that
exploit sequential structure. TS-Except-Last isolates temporal order
relative to the target frame: if accuracy remains high the model is
target-frame dominated; if it drops, the model relies on ordering.

\paragraph{(6) Frame Replacement (FR-$k$)}
\textit{Definition:} for indices $i \in \{s, s{+}k, s{+}2k, \dots\}$
(default $s=2$, $k=3$), duplicate the preceding frame:
$\mathbf{x}[:,\, i] \leftarrow \mathbf{x}[:,\, i{-}1]$.
\textit{Expectation:} reduces temporal novelty via redundant frames;
motion-sensitive models should degrade more than appearance-driven ones.

\paragraph{(7) Resolution Degradation (RD-$p$-$q$)}
\textit{Definition:} downsample the middle $p\%$ of frames by factor $q$
via average pooling, then upsample to the original resolution by
nearest-neighbor interpolation.
\textit{Expectation:} probes robustness to spatial-detail loss. Models
that exploit temporal dynamics may compensate for degraded spatial
information, particularly when the target frame is among those degraded.

\paragraph{Generalization to arbitrary target frames}
Probes (CL) and (HL) extend naturally to an arbitrary labeled
target-frame index $\tau \in \{0, \dots, T{-}1\}$:
\begin{itemize}
    \item \textbf{(C*) Hide-Target (HT-$\tau$):}
    $\mathbf{x}[:,\, \tau,\, :,\, :] = \boldsymbol{\mu}$
    \item \textbf{(D*) Target-Only (TO-$\tau$):}
    $\mathbf{x}[:,\, t \neq \tau,\, :,\, :] = \boldsymbol{\mu}$
\end{itemize}
In this work we set $\tau = T{-}1$ (last frame), consistent with the
datasets' labeling convention.

\section{Implementation Details}
\label{app:implementation}

For a given dataset, the 2D-stacked and 3D variants are trained with identical schedules, augmentation, optimizer settings, and hardware, so that performance differences reflect architecture rather than training setup; settings differ between the real-world datasets and the UCF101-24 ablation, as detailed below. All models use the Ultralytics framework with a fixed random seed (0) and SGD (learning rate 0.01, momentum 0.937, weight decay 0.0005), with the remaining optimizer hyperparameters following the Ultralytics YOLOv8 defaults.

For the two real-world datasets, we train for 100 epochs at $960{\times}960$ input and batch size 8, without data augmentation to keep the stacked-2D versus 3D comparison clean. For the UCF101-24 ablation, we train for 20 epochs at $320{\times}320$ and batch size 8, adopting the augmentation pipeline of YOWOv2~\cite{Zhiqiang2025YOWOv2} applied consistently to every frame of a clip: random scaling (range [0.8, 1.2], $p=0.5$), random crop (jitter 0.2, $p=0.5$), horizontal flip ($p=0.5$), and HSV jitter (hue 0.1, saturation 1.5, exposure 1.5, $p=0.1$). All latency and compute figures (Fig.~\ref{fig:latency_tradeoff}) are measured in FP32 on a single NVIDIA RTX 4090.

\section{Experiment-Level View of Sequence-Length Gains}
\label{app:scatter}

\begin{figure}[h]
  \centering
  \includegraphics[width=0.5\linewidth]{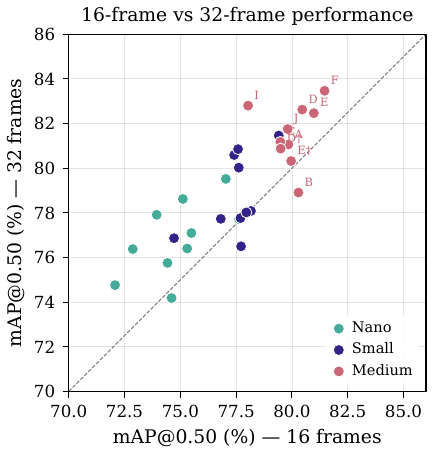}
  \caption{Scatter plot of mAP@50 at 16 frames (x-axis) versus 32 frames (y-axis) for all 30 experiments (10 configs $\times$ 3 scales). Points above the diagonal gained accuracy from longer clips. Colour encodes model scale: teal = nano, indigo = small, rose = medium. Medium-scale points are labelled with configuration IDs. The temporal-backbone configurations (D, E, F, I, J) cluster furthest above the diagonal}
  \label{fig:scatter-16v32}
\end{figure}

Figure~\ref{fig:scatter-16v32} complements the per-configuration summary
of Sec.~\ref{sec:ablation-ucf} (Fig.~\ref{fig:seq-length-gain}) with an
experiment-level scatter of 16- vs.\ 32-frame accuracy across all 30
runs, making the per-experiment spread visible behind the averaged gains.

\end{document}